\documentclass[10pt,twocolumn,letterpaper]{article}

\usepackage{cvpr}              

\usepackage{times}
\usepackage{epsfig}
\usepackage{graphicx}
\usepackage{amsmath}
\usepackage{amssymb}
\usepackage{datetime2}
\usepackage{bm}
\usepackage{color,soul}          
\usepackage{microtype}      
\usepackage{dcolumn}        
\usepackage{url}            
\usepackage{booktabs}       
\usepackage{amsfonts}       
\usepackage{nicefrac}       
\usepackage{enumitem}
\usepackage{multirow}
\usepackage{float}
\usepackage{algorithm}
\usepackage[noend]{algpseudocode}
\usepackage[numbers,sort,compress]{natbib}
\usepackage{amsthm}
\usepackage{subfiles}
\usepackage{adjustbox}
\usepackage{subcaption}
\usepackage{makecell}
\usepackage[normalem]{ulem}
\usepackage{caption}
\usepackage{nopageno}
\usepackage[flushmargin,para]{footmisc}
\usepackage[pagebackref,breaklinks,colorlinks]{hyperref} 
\usepackage[table]{xcolor}
\usepackage{pifont}
\newcommand{\cmark}{\ding{51}}%
\newcommand{\xmark}{\ding{55}}%

\usepackage[capitalize]{cleveref}
\crefname{section}{Sec.}{Secs.}
\Crefname{section}{Section}{Sections}
\Crefname{table}{Table}{Tables}
\crefname{table}{Tab.}{Tabs.}

\definecolor{lightgray}{gray}{0.97}
\definecolor{lightblue}{rgb}{0.93,0.95,1.0}

\def\cvprPaperID{276} 
\def\confName{CVPR}
\def\confYear{2022}

\begin{document}

\title{Occluded Human Mesh Recovery}

\author{
    {Rawal Khirodkar}\textsuperscript{1}\hspace{1cm} {Shashank Tripathi}\textsuperscript{2} \hspace{1cm} {Kris Kitani}\textsuperscript{1}\\
    \textsuperscript{1}Carnegie Mellon University \hspace{1cm} \textsuperscript{2}Max Planck Institute for Intelligent Systems\\
    {\urlstyle{sf} \href{https://rawalkhirodkar.github.io/ochmr}{https://rawalkhirodkar.github.io/ochmr}}
}

\twocolumn[{%
\renewcommand\twocolumn[1][]{#1}%
\maketitle
\begin{center}
    \centering
    \captionsetup{type=figure}
    \includegraphics[width=\linewidth]{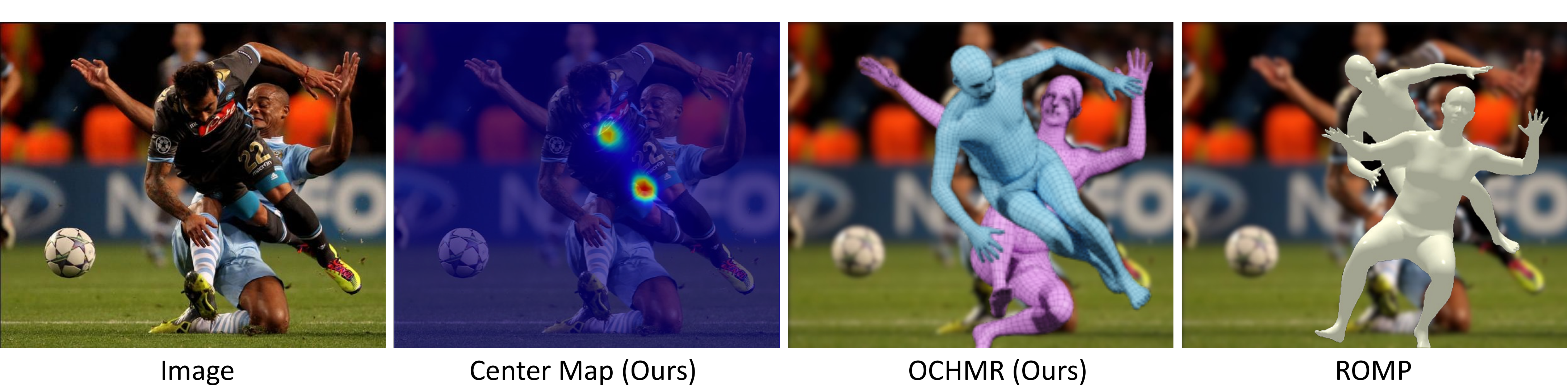}
    \caption{
    To handle severe person-person occlusion, our proposed method OCHMR conditions the deep network on image spatial context using predicted body centermaps. OCHMR is trained using multi-person mesh interpenetration and depth ordering losses. In comparison to bottom-up ROMP~\cite{sun2021monocular}, top-down OCHMR outputs pixel aligned mesh estimates for each individual under occlusion.}
    \label{figure:introduction}
\end{center}%
}]

\begin{abstract}
\vspace{-0.4cm}
Top-down methods for monocular human mesh recovery have two stages: (1) detect human bounding boxes; (2) treat each bounding box as an independent single-human mesh recovery task. Unfortunately, the single-human assumption does not hold in images with multi-human occlusion and crowding. Consequently, top-down methods have difficulties in recovering accurate 3D human meshes under severe person-person occlusion. To address this, we present Occluded Human Mesh Recovery (OCHMR) - a novel top-down mesh recovery approach that incorporates image spatial context to overcome the limitations of the single-human assumption. The approach is conceptually simple and can be applied to any existing top-down architecture. Along with the input image, we condition the top-down model on spatial context from the image in the form of body-center heatmaps. To reason from the predicted body centermaps, we introduce Contextual Normalization (CoNorm) blocks to adaptively modulate intermediate features of the top-down model. The contextual conditioning helps our model disambiguate between two severely overlapping human bounding-boxes, making it robust to multi-person occlusion. Compared with state-of-the-art methods, OCHMR achieves superior performance on challenging multi-person benchmarks like 3DPW, CrowdPose and OCHuman. Specifically, our proposed contextual reasoning architecture applied to the SPIN model with ResNet-50 backbone results in $75.2$ PMPJPE on 3DPW-PC, $23.6$ AP on CrowdPose and $37.7$ AP on OCHuman datasets, a significant improvement of $6.9$ mm, $6.4$ AP and $20.8$ AP respectively over the baseline. Code and models will be released.
\end{abstract}

\vspace{-0.6cm}
\section{Introduction}
\label{sec:intro}
Estimating accurate 3D human meshes from single images has diverse applications in modeling human-scene interactions, understanding human behaviour, AR/VR and robotics. While recent approaches~\cite{bogo2016keep,loper2015smpl,guan2009estimating,kanazawa2018end,zhang2019danet,omran2018neural,kolotouros2019learning,pavlakos2018learning} perform particularly well in images containing a single person, human mesh recovery for complex real-world scenes with multiple occluded people remains a challenging task. This can be attributed in part to simplifying assumptions made by existing methods. For instance, most top-down approaches expect a single subject in the input image, which affects robustness under in-the-wild scenarios containing severe person-person occlusion, such as crowding. In this paper, we address human mesh recovery in multi-person scenarios by mitigating the limitations of the single-person assumption of top-down approaches.

Current human mesh recovery methods can be categorized into \textit{top-down} and \textit{bottom-up} methods. Top-down methods~\cite{lin2021-mesh-graphormer,kocabas2021pare, lin2021end,li2021hybrik,guan2021bilevel,dwivedi2021learning,zhang2021pymaf,kanazawa2017end, choi2020pose2mesh} reduce the problem to a simpler task of single human mesh recovery by relying on a person detector to detect individual bounding box for each person in the image. Since each bounding box is scaled to the same size, top-down methods are less sensitive to scale variations among subjects and can achieve pixel accurate mesh alignment~\cite{zhang2021pymaf, dwivedi2021learning}. In contrast, bottom-up methods~\cite{zanfir2018deep,sun2021monocular, zhang2021body} simultaneously predict meshes for all subjects in the input image but are limited to a fixed input resolution due to computational constraints. \eg ROMP~\cite{sun2021monocular}, a bottom-up method, recovers a limited number of human meshes from a resized $512 \times 512$ input whereas SPIN~\cite{kolotouros2019learning}, a top-down method, scales each bounding box to $224 \times 224$, retaining higher input resolution per person(see Fig. \ref{figure:introduction}). This observation has also been discussed by Cheng~\etal~\cite{cheng2020higherhrnet} albeit in the context of 2D human pose estimation. Thus, top-down methods are currently the best performers on various multi-human benchmarks~\cite{ionescu2013human3, von2018recovering, mehta2017monocular,joo2015panoptic,joo2018total,sigal2010humaneva}. Despite the advantages, due to the single-human assumption, when presented with multi-human inputs like crowded scenes, top-down methods are forced to select a single plausible mesh per detection bounding box. Bottom-up methods do not have this limitation and typically perform better under occlusion.

\begin{figure}
\centering
\includegraphics[width=1\linewidth, height=0.95\linewidth]{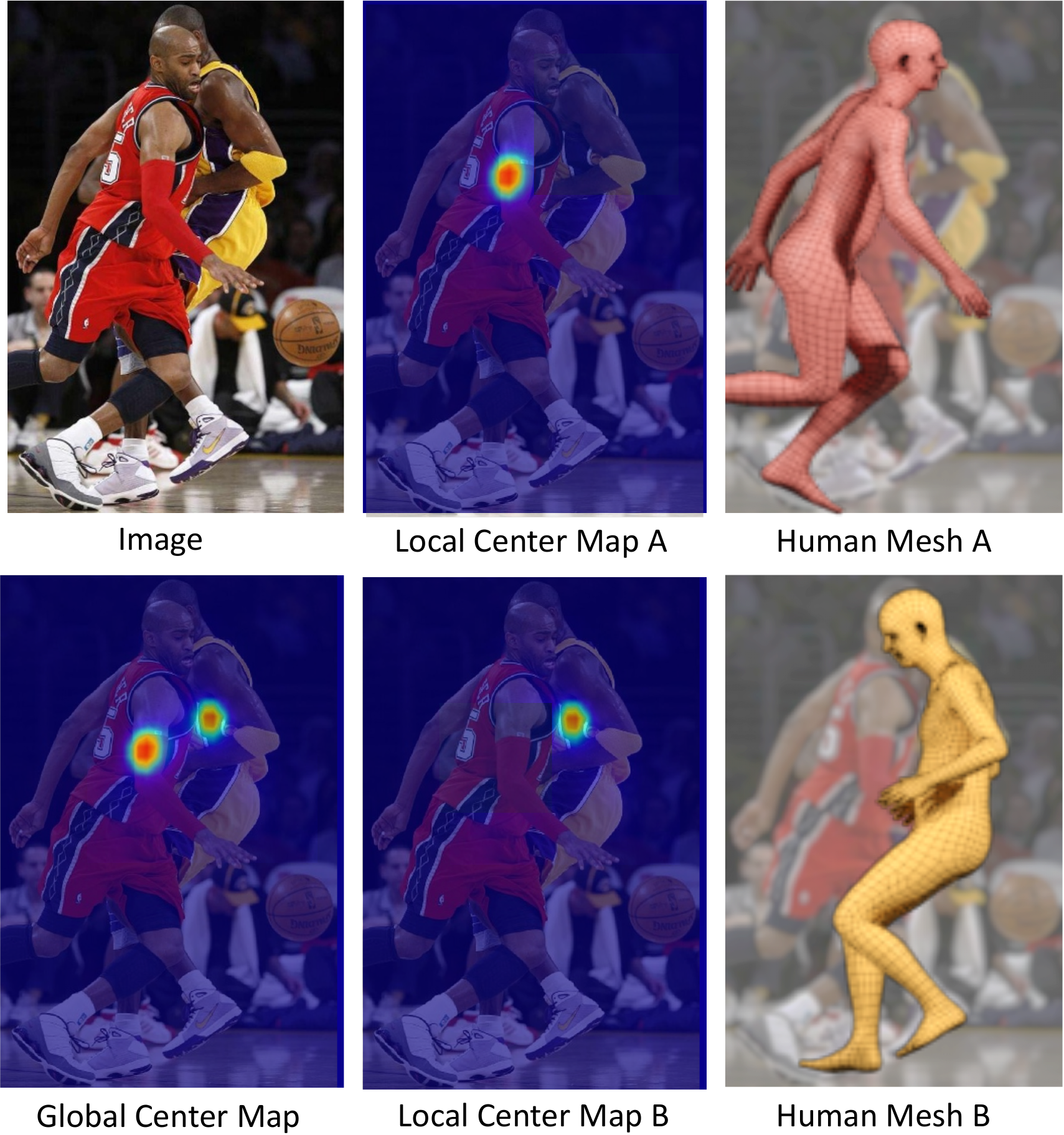}
\caption{OCHMR leverages image spatial-context for occlusion reasoning by predicting body centermaps. The deep network predicts the mesh output using input image, the subject-specific local centermap and the scene-specific global centermap.}
\label{figure:context}
\vspace*{-0.2in}
\end{figure}

A general method should have both traits -- be robust to scale variations and person-person occlusions. To this end, we rethink top-down human mesh recovery by predicting \textit{multiple} meshes from the input bounding box. We condition the top-down model on image \textit{spatial-context} in the form of body-center maps, refer Fig. \ref{figure:context}. Our choice of using center maps for representing humans under occlusion is inspired by crowd-counting literature~\cite{song2021rethinking, ma2021towards, wang2021uniformity} and recent works in detection~\cite{duan2019centernet,zhou2019bottom,zhou2019objects}. Our method, OCHMR, predicts the output mesh from the input image for the person of interest in the subject-specific \textit{local} center-map. Similar to bottom-up methods, we also use information from the \textit{global} center-map for understanding overall scene context, which is helpful for occlusion reasoning. With this strategy, we obtain the best of both worlds -- OCHMR achieves pixel accurate mesh alignment similar to top-down methods and is robust to occlusions similar to bottom-up methods (See Fig. \ref{figure:introduction}).

To design a top-down architecture capable of contextual conditioning using centermaps, we adopt the mechanism of feature normalization~\cite{perez2018film, dumoulin2016learned, huang2017arbitrary} and propose a novel Context Normalization (CoNorm) block to process the global and local centermaps. The CoNorm blocks are used to inject contextual information at multiple depths in the deep feature backbone network. The spatial context is necessary for 3D occlusion reasoning, and the CoNorm block allows for adaptive normalization of intermediate features of the network without changing the backbone. We show that unlike \textit{early fusion} (\eg channel-wise concatenation) of centermaps with input image $\mathbf{I}$, CoNorm can effectively utilize the contextual information from the image. OCHMR is general and can be extended to other top-down human mesh recovery methods with minimal effort. 

While the use of spatial-context allows our method to reason about occlusions, our method must also reason about the intersection of a set of 3D human meshes. To address this, following CRMH~\cite{jiang2020coherent}, we use an interpenetration loss to penalize intersections among reconstructed meshes and a differentiable depth-ordering loss for depth-consistent human mesh recovery. Furthermore, we make use of training-time data augmentation like scaling and cropping, which affords OCHMR the ability to predict meshes from a variety of body-center locations. We show that our proposed method is also robust to errors in estimated body-centers under severe-occlusion. Our empirical results show that OCHMR does not require precise centermaps that correspond to actual body-centers but can also work with any point in its vicinity.

Overall, OCHMR outperforms both top-down and bottom-up methods on various datasets. For challenging datasets such as 3DPW-PC~\cite{zhang2020object}, CrowdPose~\cite{li2019crowdpose} and OCHuman~\cite{zhang2019pose2seg}, containing a larger proportion of cluttered scenes (with multiple overlapping people), OCHMR sets a new state-of-the-art for 3D reconstruction error (PMPJPE) and 2D keypoint average precision (AP) achieving $77.1$ PMPJPE, $21.4$ AP and $24.8$ AP respectively on the \texttt{val} sets outperforming bottom-up methods (Tab.~\ref{table:occlusion}). Further, when evaluating using ground-truth bounding boxes, OCHMR dramatically improves SPIN~\cite{kolotouros2019learning} by $20.8$ AP and $6.4$ AP on the OCHuman and CrowdPose dataset respectively. In summary:
\begin{itemize}
    \item OCHMR advances top-down human mesh recovery methods by addressing limitations caused by the single-human assumption. Our method leverages spatial-context in the form of centermaps to predict multiple mesh outputs from an input image.
    \item We introduce novel Context Normalization (CoNorm) blocks to inject global and local centermap information at multiple depths of the top-down network.
    \item Our approach achieves state-of-the-art results on the occluded 3DPW-PC, CrowdPose and OCHuman datasets. Empirically, we also show that OCHMR is resilient to noisy body center estimates and demonstrates robust 3D reasoning using multi-person losses. 
\end{itemize}

\section{Related Work}
\label{sec:related_work}
Deep learning has significantly advanced 3D human mesh recovery~\cite{kanazawa2017end, kolotouros2019learning, zhang2021pymaf,lin2021end,li2021hybrik,kissos2020beyond,kocabas2020vibe, lin2021-mesh-graphormer, tripathi2020posenet3d, kocabas2021pare, guan2021bilevel, dwivedi2021learning, choi2020pose2mesh}, facilitating the more challenging task of mesh recovery under severe multi-person occlusion~\cite{zhang2019pose2seg,li2019crowdpose,zhang2020object,sun2021monocular,jiang2020coherent}, which is the main focus of this work.

\textbf{Biased Human Mesh Recovery Benchmarks.} Most benchmark datasets~\cite{ionescu2013human3, von2018recovering, mehta2017monocular,joo2015panoptic,joo2018total,sigal2010humaneva, yi2022mover} used for learning human mesh recovery focus on a single person and do not accurately represent the distribution of possible occlusions present in the real world. Human3.6M~\cite{ionescu2013human3}, HumanEva~\cite{sigal2010humaneva} and TotalCapture~\cite{joo2018total} are popular datasets collected using motion capture (mocap) systems using optical markers. While providing accurate annotations, they only have a single subject in the image with limited image complexity due to the lack of background variation. In contrast, datasets like MPI-INF-3DHP~\cite{mehta2017monocular}, PanopticStudio~\cite{joo2015panoptic} and 3DPW~\cite{von2018recovering} contain multi-person annotations but have limited person-person occlusion -- less than 27\% of all annotations have crowding (at IoU 0.5). Although previous methods~\cite{kanazawa2017end, kolotouros2019learning, zhang2021pymaf,lin2021end,li2021hybrik,kissos2020beyond,kocabas2020vibe} leverage 2D keypoint annotations from datasets like COCO~\cite{lin2014microsoft}, MPII~\cite{andriluka20142d}, LSP-Extended~\cite{johnson2011learning}, the 2D datasets are also known to contain similar biases~\cite{ruggero2017benchmarking, zhang2019pose2seg, khirodkar2021multi}. These biases have affected critical design decisions in state-of-the-art methods which lead to poor generalization under heavy occlusion~\cite{sun2021monocular, jiang2020coherent}. Recently, challenging datasets such as OCHuman~\cite{zhang2019pose2seg}, CrowdPose~\cite{li2019crowdpose} and 3DPW-PC~\cite{zhang2020object} containing heavy occlusion have been proposed to capture these biases. OCHMR shows a significant improvement over existing works under such challenging conditions.

\textbf{Top-Down Human Mesh Recovery.} Top-down methods~\cite{lin2021-mesh-graphormer,kocabas2021pare, lin2021end,li2021hybrik,guan2021bilevel,dwivedi2021learning,zhang2021pymaf,kanazawa2017end, choi2020pose2mesh} estimate 3D human mesh of a single person
within a person bounding box. The bounding box is usually generated using a person detector~\cite{cheng2018revisiting,lin2017feature,liu2016ssd,bochkovskiy2020yolov4,ren2015faster,he2017mask}. As the input bounding boxes are cropped and scaled to the same size, top-down methods are less sensitive to person scale variations in the image. In contrast, bottom-up methods have to deal with scale variations which compromises pixel alignment in the reconstruction results. For these reasons~\cite{cheng2020higherhrnet}, most state-of-the-art 2D pose estimation methods~\cite{liu2021polarized, zhang2020distribution, sun2019deep, mcnally2021evopose2d} are also top-down. However, top-down methods inherently assume a single person in the input image and often fail under occlusions in multi-person scenarios.  Recent works like ~\cite{moon2020i2l, tripathi2020posenet3d,choi2020pose2mesh,sengupta2020synthetic,joo2020exemplar,bogo2016keep} use 2D/3D poses as input along with bounding boxes for human mesh recovery. However, obtaining accurate 2D poses under occlusion is difficult and pose errors like joint swaps~\cite{ruggero2017benchmarking} are magnified during the 3D reconstruction~\cite{joo2020exemplar}. CRMH~\cite{jiang2020coherent} handles multi-person scenarios by using RoI-aligned~\cite{he2017mask} features of each person to predict the SMPL~\cite{loper2015smpl} parameters. However, the reliance on bounding-box-level features makes it hard to effectively differentiate between two overlapping bounding boxes. OCHMR resolves these issues by conditioning the top-down model on image context in the form of \textit{body-centers} -- a representation which helps in resolving ambiguity under multi-person occlusion.

\begin{figure*}[t]
\centering
\includegraphics[width=1\linewidth]{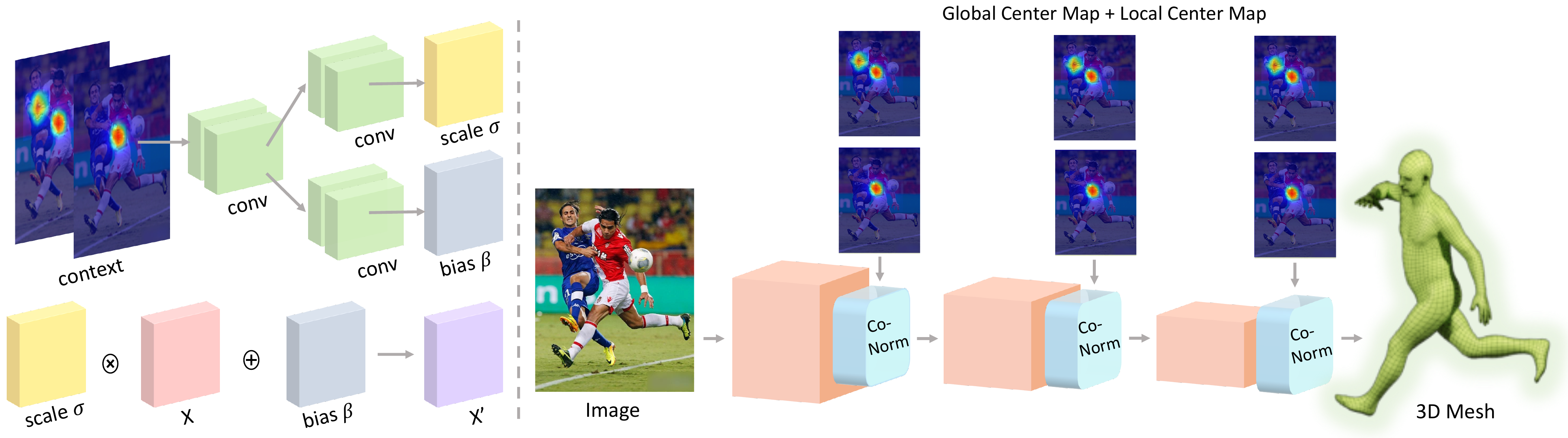}
\caption{Context Normalization (CoNorm) Block (in \textcolor{cyan}{blue}) learns scale $\boldsymbol{\sigma}$ and bias $\boldsymbol{\beta}$ parameters for spatial affine transformation of intermediate features $\mathbf{X}$ (in \textcolor{magenta}{red}) from the image context. \textit{Left}: We concatenate 2D global and local center maps channel-wise to represent image context. \textit{Right}: We insert multiple CoNorm blocks at various depths in the deep neural architecture - injection of high resolution contextual information throughout the network is critical for predicting accurate 3D meshes under occlusion. }
\label{figure:method_architecture}
\vspace*{-0.2in}
\end{figure*}


\textbf{Bottom-Up Human Mesh Recovery.} Unlike top-down methods, only few methods exist which use the bottom-up paradigm for human mesh recovery. Zanfir \textit{et al.}~\cite{zanfir2018deep} uses intermediate 3D poses to estimate the 3D mesh of each person in a bottom-up fashion. ROMP~\cite{sun2021monocular} uses a fixed resolution body-center map to disambiguate between multiple persons under occlusion. Due to the fixed input size, $512\times 512$, ROMP is limited to predicting a small numbers of meshes. In contrast, OCHMR is top-down and can leverage input resizing of subject bounding box to a higher resolution for pixel accurate shape estimation. Being top-down, OCHMR can be applied to all detected persons in the input image.

\section{Method}
\label{sec:method}
OCHMR leverages the strengths of both top-down and bottom-up methods for multi-person mesh recovery under severe person-person occlusion/crowding. In this section, we briefly describe the top-down method used as baseline architecture in our approach. Then we provide details of our contextual representations i.e. local and global centermaps and the context estimation network. Finally, we describe the proposed architectural improvements in the form of Context Normalization (CoNorm) blocks and multi-person losses used in training.

\textbf{Top-down Human Mesh Recovery.} Top-down human mesh recovery aims to predict a 3D human mesh from an input image $\mathbf{I} \in \mathbb{R}^{H \times W \times 3}$. Most top-down methods transform this problem to estimating the parameters of a human body model like SMPL~\cite{loper2015smpl} and the camera parameters. We represent body pose, shape and camera parameters by $\boldsymbol{\Theta} = [\boldsymbol{\theta_{\text{pose}}}, \boldsymbol{\theta_{\text{shape}}}, \boldsymbol{\theta_{\text{camera}}}]$, $\boldsymbol{\theta_{\text{pose}}} \in \mathbb{R}^{24 \times 6}, \boldsymbol{\theta_{\text{shape}}} \in \mathbb{R}^{10}, \boldsymbol{\theta_{\text{camera}}} \in \mathbb{R}^{3}$. The pose parameters $\boldsymbol{\theta_{\text{pose}}}$ are the 6D representation of the joint rotations~\cite{zhou2019continuity} of the 24 body joints and include the global root orientation of the SMPL body. The shape parameters $\boldsymbol{\theta_{\text{shape}}}$ represent the first 10 coefficients of the PCA shape space. The camera parameters $\boldsymbol{\theta_{\text{camera}}}$ describe the 2D scale $s$ and translation $\mathbf{t} = (t_x, t_y)$. SMPL is linear and fully differentiable, making it a suitable representation for learning based methods.

Similar to~\cite{kolotouros2019learning, kanazawa2019learning}, we define a deep regression model $P$ as our baseline top-down architecture for human mesh recovery. The bounding box at training and inference is scaled to $H \times W$ and is provided as an input to $P$. Let $\boldsymbol{\Theta}^{\text{gt}}$ denote the ground-truth SMPL and camera parameters corresponding to the human in the input image $\mathbf{I}$. The deep regression model $P$ transforms input $\mathbf{I}$ to a single 3D mesh $\boldsymbol{M}$, such that $\boldsymbol{\Theta} = P(\mathbf{I})$. $P$ is trained to minimize the sum of various 2D/3D pose and shape losses (using 2D pose annotations and segmentations masks if available) denoted by $\mathcal{L}(\boldsymbol{\Theta}^{\text{gt}}, \boldsymbol{\Theta})$~\cite{kolotouros2019learning}.

\vspace*{-0.05in}
\subsection{Occluded Human Mesh Recovery}
We propose to modify the top-down deep regression model $P$ to predict multiple meshes as follows. Let $N$ be the number of ground-truth subjects present in the image $\mathbf{I}$. $N$ is set to the total number of subjects with atleast $5$ visible 2D keypoints in the image. Let $\boldsymbol{\Theta}^{\text{gt}}_0, \boldsymbol{\Theta}^{\text{gt}}_1, \dots, \boldsymbol{\Theta}^{\text{gt}}_{N-1}$ be the corresponding ground-truth mesh parameters. Our modified deep regression model $P$ predicts $N$ instances, $\boldsymbol{\Theta}_0, \boldsymbol{\Theta}_1, \dots, \boldsymbol{\Theta}_{N-1}$ for an input $\mathbf{I}$. This is achieved by conditioning the network $P$ on the \textit{spatial-context} $\mathbf{C}_i$ individually for each subject. $P$ accepts both $\mathbf{I}$ and $\mathbf{C}_i$ as input and predicts $\boldsymbol{\Theta}_i = P(\mathbf{I}, \mathbf{C}_i)$ where $i \in \{0, 1, \dots, N-1 \}$. We define the OCHMR's single person loss $\mathcal{L}_{\text{single}}$ as follows, 

\vspace*{-0.1in}
\begin{eqnarray}
    \mathcal{L}_{\text{single}}& = &\frac{1}{N} \sum_{i=0}^{N-1} \mathcal{L}(\boldsymbol{\Theta}^{\text{gt}}_i, \boldsymbol{\Theta}_i)
\end{eqnarray}

During inference, we vary the spatial-context $\mathbf{C}_i$ to extract multiple mesh predictions from the same input image $\mathbf{I}$. In cases of severely overlapping people, it is hard for the baseline top-down method to estimate diverse body meshes $\mathbf{M}$ from similar image patches $\mathbf{I}$. OCHMR uses spatial-context $\mathbf{C}$ to resolve the implicit ambiguity of the bounding-box input representation in such multi-person cases.

\subsection{Global and Local Center Map Estimation}
Our top-down framework relies heavily on the representation of the spatial-context $\mathbf{C}$. It is crucial to define a representation which is explicit and robust to occlusion. Inspired by ~\cite{duan2019centernet, sun2021monocular}, we choose body centers to encode the spatial context $\mathbf{C}$ of the image. Specifically, we represent the contextual information of $i^{\text{th}}$ instance as $\mathbf{C}_i = (\mathbf{C}_{\text{global}}, \mathbf{C}_{\text{local-i}})$ where $\mathbf{C}_{\text{global}}$ is the body-center heatmap of all the $N$ instances present in the image $\mathbf{I}$ and $\mathbf{C}_{\text{local-i}}$ is the body-center heatmap of the $i^{\text{th}}$ instance (see Fig.~\ref{figure:context}). $\mathbf{C}_{\text{local-i}}$ is calculated by thresholding and iterating over pixel locations in $\mathbf{C}_{\text{global}}$. While $\mathbf{C}_{\text{local-i}}$ informs the network about the subject of interest, $\mathbf{C}_{\text{global}}$ places the subject in the context of its neighbors, thereby helping the network disambiguate between occluding persons. 

The body center is defined as the center of visible torso joints (neck, left/right shoulders, pelvis, and left/right hips). When all torso joints are invisible, the center is the average of the visible joints. Following~\cite{sun2021monocular}, we calculate the ground-truth body center from the ground-truth 2D pose. All the ground-truth 2D body center locations are converted into $\mathbf{C}^{\text{gt}}_{\text{global}}$ which is a heatmap of size $H \times W$ indicating the probability of the body centers at any spatial location~\cite{sun2019deep}. At inference, we use a fully-convolutional~\cite{long2015fully} neural network $F$ to predict $\mathbf{C}_{\text{global}}$ from the input image $\mathbf{I}$.  $F$ is trained to minimize the mean squared loss $\mathcal{L}_{\text{context}} = \mathtt{MSE}(\mathbf{C}^{\text{gt}}_{\text{global}}, \mathbf{C}_{\text{global}})$. Finally, the context of the $i^{\text{th}}$ instance $\mathbf{C}_i$ is the channel-wise concatenation of $\mathbf{C}_{\text{global}}$ and $\mathbf{C}_{\text{local-i}}$ \ie $\mathbf{C}_i \in \mathbb{R}^{H \times W \times 2}$.

\subsection{Context Normalization Block}
A key challenge is to design an architecture that incorporates spatial-context as a conditioning input. A na\"{i}ve \textit{early fusion} approach would be to simply concatenate the input image $\mathbf{I}$ with the spatial-context $\mathbf{C}$. Similarly, \textit{late fusion} would concatenate feature maps from later layers within the network with appropriately down-sampled context $\mathbf{C}$. However, both of these approaches fail to improve performance. 

We describe the Context Normalization (CoNorm) block that can be easily introduced in any existing feature extraction backbone to overcome this issue (see \cref{figure:method_architecture}). The key intuition is that CoNorm allows normalization of intermediate feature maps using the conditioning input $\mathbf{C}$. 
The deep regression model $P$ uses CoNorm blocks to leverage contextual information for predicting multiple meshes from the input image $\mathbf{I}$. Similar to Batch Normalization~\cite{ioffe2015batch}, CoNorm learns to influence the output of the neural network by applying an affine transformation to the network's intermediate features based on $\mathbf{C}$.

Let $\mathbf{X} \in \mathbb{R}^{H' \times W' \times D}$ be an intermediate feature in the deep network $P$. The CoNorm block consists of operations $\mathbf{\Phi}_{\text{latent}}$, $\mathbf{\Phi}_{\text{scale}}$ and $\mathbf{\Phi}_{\text{bias}}$ on the context $\mathbf{C}$. $\mathbf{C}$ is spatially downsampled to the same 2D resolution $H' \times W'$ as $\mathbf{X}$. 
\vspace*{-0.05in}
\begin{eqnarray}
  \boldsymbol{\lambda} & = & \mathbf{\Phi}_{\text{latent}}(\mathbf{C}), \\
  \boldsymbol{\sigma} & = & \mathbf{\Phi}_{\text{scale}}(\boldsymbol{\lambda}), \\
  \boldsymbol{\beta} & = & \mathbf{\Phi}_{\text{bias}}(\boldsymbol{\lambda}), \\
 \mathbf{X}' & = & \boldsymbol{\sigma}*\mathbf{X} + \boldsymbol{\beta}.
\end{eqnarray}

$\mathbf{\Phi}_{\text{latent}}$ maps $\mathbf{C}_i$ to $\boldsymbol{\lambda}$ which is in a $V$ dimensional latent space \ie $\boldsymbol{\lambda} \in \mathbb{R}^{H' \times W' \times V}$. $\mathbf{\Phi}_{\text{scale}}$ and $\mathbf{\Phi}_{\text{bias}}$ use the latent vector $\boldsymbol{\lambda}$ to predict $\boldsymbol{\sigma}$ and $\boldsymbol{\beta}$ respectively.  $\boldsymbol{\sigma}, \boldsymbol{\beta} \in \mathbb{R}^{H' \times W' \times D}$. We use the predicted $\boldsymbol{\sigma}$ and $\boldsymbol{\beta}$ to normalize the intermediate feature $\mathbf{X}$ using element-wise operations to output $\mathbf{X'}$.


\subsection{Multi-Person Losses}
\begin{figure}[b]
\centering
\includegraphics[width=0.7\linewidth]{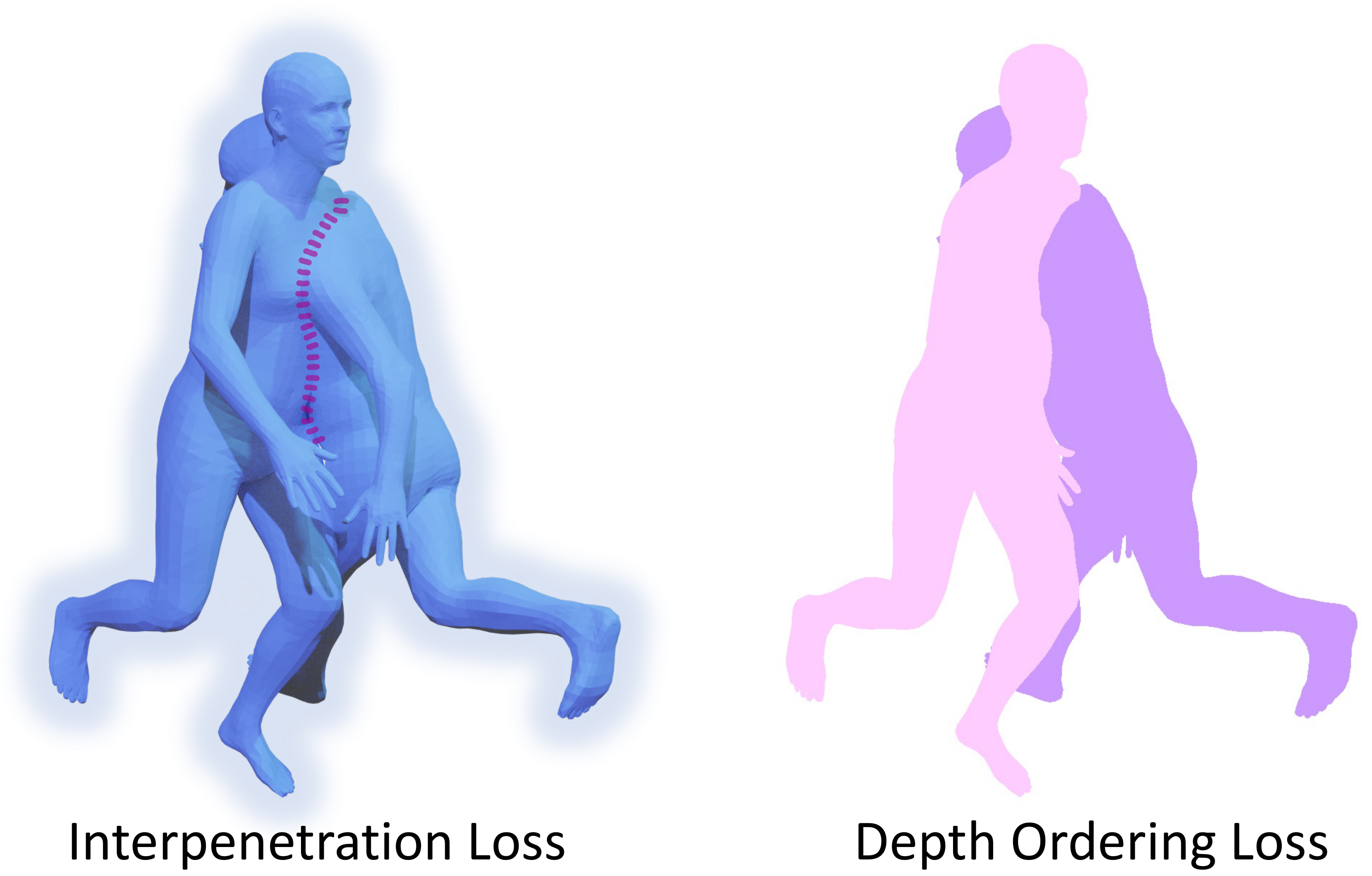}
\caption{Interpenetration loss prevents mesh intersections. Depth ordering loss is useful for depth-consistent mesh outputs.}
\vspace*{-0.1in}
\label{figure:method_losses}
\end{figure}

In multi-person scenarios, the regression model $P$ can often predict meshes that are intersecting and have incoherent depth ordering. Following~\cite{jiang2020coherent}, we adopt two multi-person losses -- i) interpenetration and ii) depth-ordering loss, refer Fig.~\ref{figure:method_losses}. We briefly describe the losses here for completeness but refer to~\cite{jiang2020coherent} for more details.

\begin{table*}[t]
    \centering
    \small
    \rowcolors{3}{}{lightgray}
    \renewcommand{\arraystretch}{1.2} 
    \setlength{\tabcolsep}{4pt}
    \begin{tabular}{@{}l|c|l c c| l c c c c | l c c @{}}
    \Xhline{3\arrayrulewidth}
   
    \multirow{2}{*}{\textbf{Method}} & \textbf{Extra} & \multicolumn{3}{c|}{\textbf{3DPW-PC} $\downarrow$} & \multicolumn{5}{c|}{\textbf{OCHuman$\uparrow$}} & \multicolumn{3}{c}{\textbf{CrowdPose$\uparrow$}}   \\
    
     & \textbf{Data} & \textbf{MPJPE} & \textbf{PMPJPE} & \textbf{PVE} & \textbf{$\text{AP}$} & \textbf{$\text{AP}^{50}$} & $\text{AP}^{75}$ & \textbf{$\text{AP}^\text{M}$} & \textbf{$\text{AP}^\text{L}$} & \textbf{$\text{AP}$} & \textbf{$\text{AP}^{50}$} & \textbf{$\text{AP}^{75}$}\\
    \hline
    SPIN~\cite{kolotouros2019learning} & \cmark& 129.6 & 82.6 & 157.6 & 12.7 & 46.8 & 19.4 & 17.8 & 26.2 & 16.4 & 40.1 & 10.6 \\
    PyMaf~\cite{zhang2021pymaf}& \cmark & 126.7 & 81.3 & 154.3 & 14.3 & 48.7 & 21.5 & 18.0 & 28.7 & 17.4 & 42.7 & 13.0 \\
    ROMP$\star$~\cite{sun2021monocular} & \cmark& 115.6 & 75.8 & 147.5 & 19.8 & 56.2 & 25.0 & 19.3 & 32.9 & 28.5 & 58.8 & 24.7 \\ 
    \hline
    SPIN~\cite{kolotouros2019learning}& \xmark& 132.7 & 83.7 & 162.3 & 11.1 & 41.4 & 18.6 & 15.6 & 25.9 & 14.8 & 38.5 & 9.5 \\
    ROMP~\cite{sun2021monocular} & \xmark& 119.7 & 79.7 & 152.8 & 15.6 & 55.0 & 23.6 & 18.7 & 30.0 & 18.9 & 44.6 & 13.8 \\
    OCHMR (Ours) & \xmark& \textbf{117.5 (-2.2)} & \textbf{77.1} & \textbf{149.6} & \textbf{24.8 (+9.2)} & \textbf{60.7} & \textbf{28.6} & \textbf{22.3} & \textbf{34.2} & \textbf{21.4 (+2.5)} & \textbf{48.3} & \textbf{16.5} \\
    \hline
    \multicolumn{12}{c}{Using ground-truth bounding boxes} \\
    \hline
    SPIN~\cite{kolotouros2019learning} &\xmark & 128.4 & 82.1 & 155.7 & 16.9 & 56.1 & 25.4 & 20.0 & 31.4 & 17.2 & 42.4 & 11.2 \\
    OCHMR (Ours) & \xmark& \textbf{112.2 (-16.2)} & \textbf{75.2} & \textbf{145.9} & \textbf{37.7 (+20.8)} & \textbf{76.4} & \textbf{33.0} & \textbf{25.0} & \textbf{37.7} & \textbf{23.6 (+6.4)} & \textbf{51.1} & \textbf{18.7}\\
    \Xhline{3\arrayrulewidth}
    \end{tabular}
    \caption{Comparisons to the state-of-the-art methods under severe occlusion using FasterRCNN~\cite{ren2015faster} and ground-truth bounding boxes. OCHMR significantly outperforms top-down as well as bottom-up approaches across all benchmarks. ROMP$\star$ trains on CrowdPose.}
    \label{table:occlusion}
\vspace*{-0.2in}
\end{table*}


\textbf{Interpenetration Loss.}
Let $\mathbf{\Omega}$ be the modified Signed Distance Field (SDF)~\cite{hassan2019resolving} over the 3D space. $\mathbf{\Omega}$ takes a positive value for all the points inside the 3D human mesh $\mathbf{M}$, proportional to the distance from the mesh surface and is $0$ everywhere else. We compute a separate distance field $\mathbf{\Omega}_i$ for each human mesh $ \mathbf{M}_i \in \{0, 1, \dots, N\}$ in the image $\mathbf{I}$. We define the pairwise interpenetration loss $\mathcal{L}^{ij}_{\text{collision}}$ between mesh $\mathbf{M}_i$ and mesh $\mathbf{M}_j$ as follows,
\vspace*{-0.05in}
\begin{eqnarray}
    \mathbf{\Omega}(x, y, z) &=& -\text{min}(\text{SDF}(x,y,z), 0), \\
    \mathcal{L}^{ij}_{\text{collision}} &=& \sum_{v \in \mathbf{M}_j} \mathbf{\Omega}_i(v), \\
    \mathcal{L}_{\text{collision}} &=& \mathop{\sum_{i=1}^{N}\sum_{j=1}^{N}}_{i \neq j} \mathcal{L}^{ij}_{\text{collision}}.
\end{eqnarray}

\noindent
$\mathcal{L}_{\text{collision}}$ is the sum of valid pairwise mesh collisions (Fig.\ref{figure:method_losses}).

\textbf{Depth-ordering Loss.}
We now define the depth-ordering loss $\mathcal{L}_{\text{depth}}$. The key idea is to leverage the ground-truth instance segmentation maps available in the COCO datasets~\cite{lin2014microsoft}. We render all the meshes and the corresponding depth maps onto the image plane using a differentiable renderer~\cite{kato2018neural} and optimize the vertex locations based on the agreement with the ground-truth instance segmentation map of the image $\mathbf{I}$ (Fig.\ref{figure:method_losses}).

Finally, we train the network $P$ to minimize the loss $\mathcal{L}$ where $w_{\text{single}}, w_{\text{collision}}$ and $w_{\text{depth}}$ are loss weights,
\begin{eqnarray}
    \mathcal{L} = w_{\text{single}}\mathcal{L}_{\text{single}} + w_{\text{collision}}\mathcal{L}_{\text{collision}} + w_{\text{depth}}\mathcal{L}_{\text{depth}}
\end{eqnarray}

\section{Experiments}
\label{sec:experiments}
\subsection{Implementation Details}

\textbf{OCHMR.} For a fair comparison with other approaches~\cite{sun2021monocular, kolotouros2019learning}, we use ResNet-50~\cite{he2016deep} as the default backbone for the mesh regression model $P$ and HRNet-W32~\cite{sun2019deep} as the backbone for the context estimator $F$. We insert CoNorm blocks after each of the $4$ ResNet block in the backbone. We set the CoNorm's latent space dimensionality $K$ as $128$ for all our experiments. The input images are resized to $224 \times 224$, keeping the same aspect ratio and padding with zeros. Following~\cite{sun2019deep}, gaussians of size $6$ pixels is used to generate the local/global centermaps. The train-time data-augmentation, training schedule and all other hyper-parameters are set similar to ~\cite{kolotouros2019learning}. The loss weights are set to $w_{\text{single}}=1$, $w_{\text{collision}}=0.2$, $w_{\text{depth}}=0.4$ to ensure that the weighted loss items are of the same magnitude. The threshold of the local/global center heatmaps is set to $0.3$.


\textbf{Training Datasets.} Similar to~\cite{kolotouros2019learning}, we use MPI-INF-3DHP~\cite{mehta2017monocular}, COCO~\cite{lin2014microsoft}, MPII~\cite{andriluka20142d}, LSP-Extended~\cite{johnson2011learning} for training (we do not use Human3.6M~\cite{ionescu2013human3} due to licensing issues). Only the training sets are used, following the standard split protocols. We use ground-truth SMPL annotations from MPI-INF-3DHP and 2D annotations from COCO, MPII and LSP-Extended. The instance segmentation masks from COCO are used to compute $\mathcal{L}_{\text{depth}}$. 

\textbf{Evaluation Benchmarks.} 3DPW-PC~\cite{zhang2020object} is employed as the main benchmark for evaluating 3D mesh/joint error since it contains in-the-wild multi-person videos with abundant 2D/3D annotations. 3DPW-PC is the \textit{person-occluded} subset of 3DPW~\cite{von2018recovering}. We also evaluate OCHMR under severe occlusion on Crowdpose~\cite{li2019crowdpose} and OCHuman~\cite{zhang2019pose2seg} which are crowded-in-the-wild 2D pose benchmarks. For completeness, we also benchmark our approach on the general datasets like 3DPW and COCO.

\textbf{Evaluation Metrics.} We report mean per joint position error (MPJPE), Procrustes-aligned MPJPE (PMPJPE) and per-vertex error (PVE) on the 3D datasets. MPJPE and PMPJPE evaluates the 3D joint rotation accuracy and PVE evaluates the 3D surface error. Also, to evaluate the pose accuracy under occlusion, we report standard metrics such as $\text{AP}, \text{AP}^{50}, \text{AP}^{75}, \text{AP}^\text{M}, \text{AP}^\text{L}, \text{AR}$ at various Object Keypoint Similarity~\cite{lin2014microsoft,li2019crowdpose}. We also report results using bounding boxes obtained via Faster R-CNN~\cite{ren2015faster} detector.

\subsection{Comparison to the State-of-the-Art}
\begin{figure*}
 \captionsetup{font=small}
 \begin{center}

\includegraphics[height=0.16\textheight,width=0.48\linewidth]{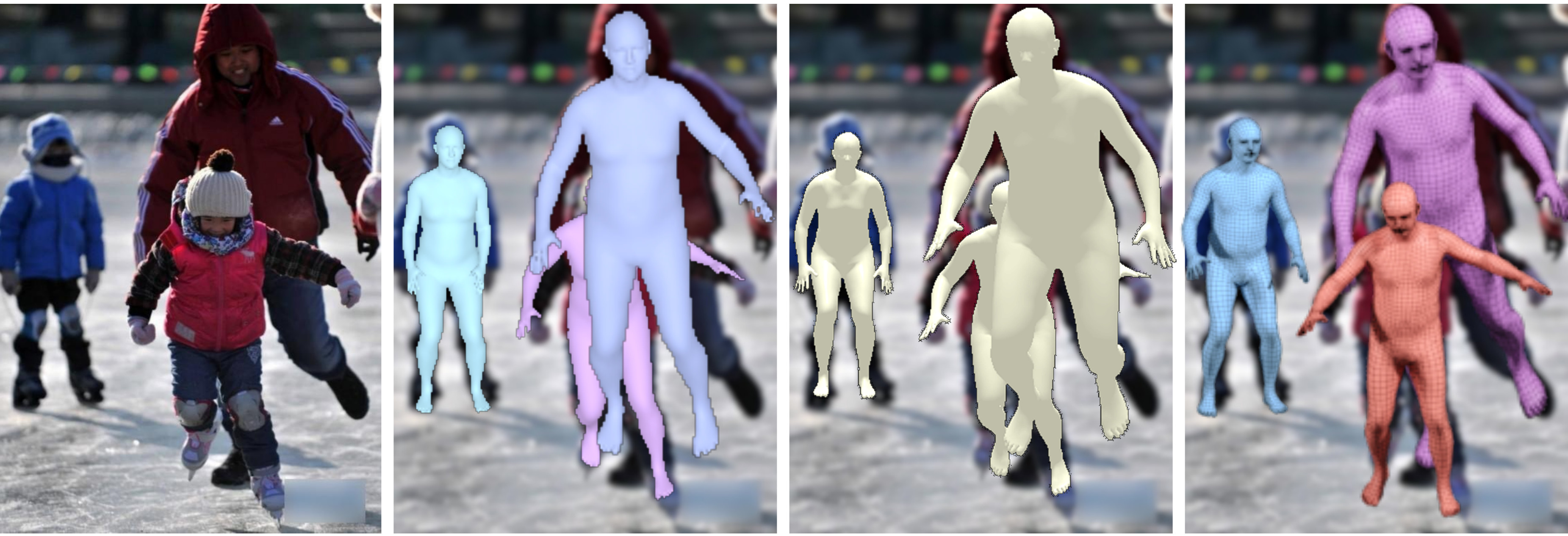}
\includegraphics[height=0.16\textheight,width=0.48\linewidth]{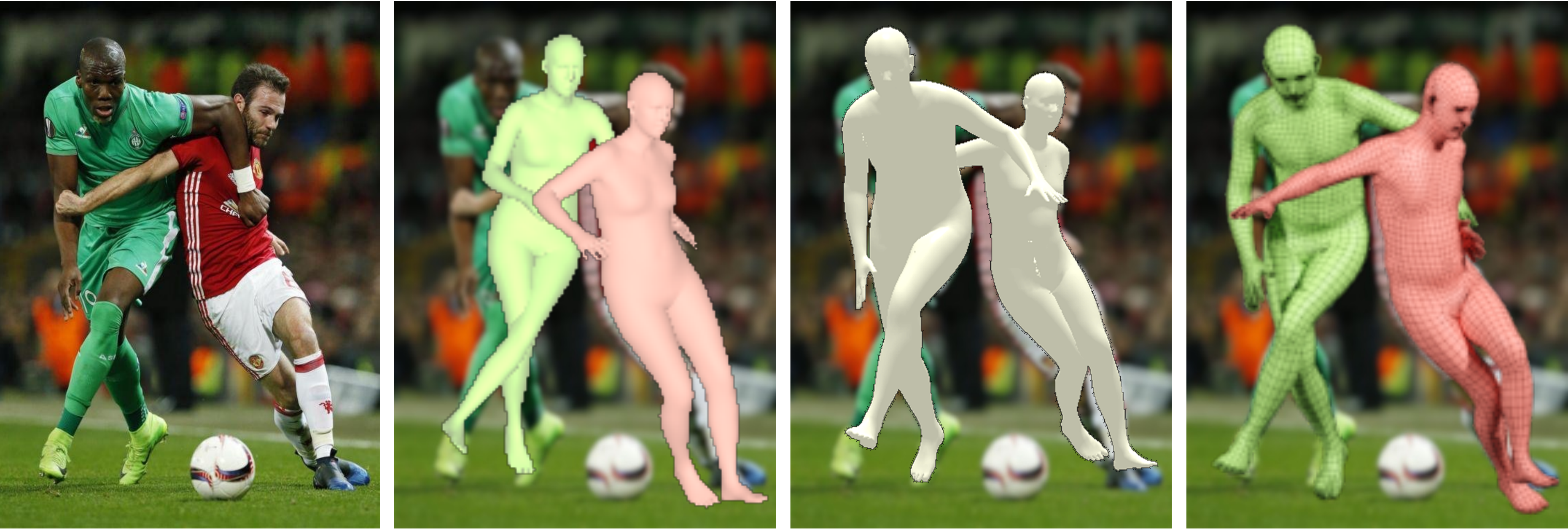}

\includegraphics[height=0.16\textheight,width=0.48\linewidth]{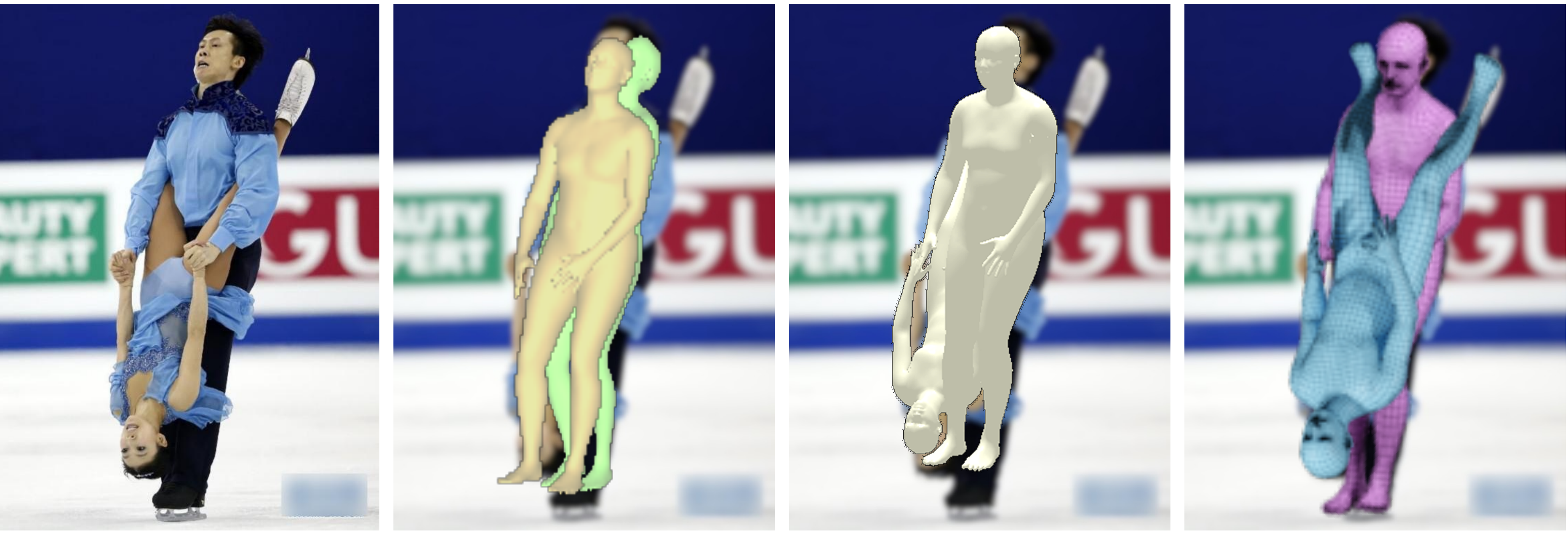}
\includegraphics[height=0.16\textheight,width=0.48\linewidth]{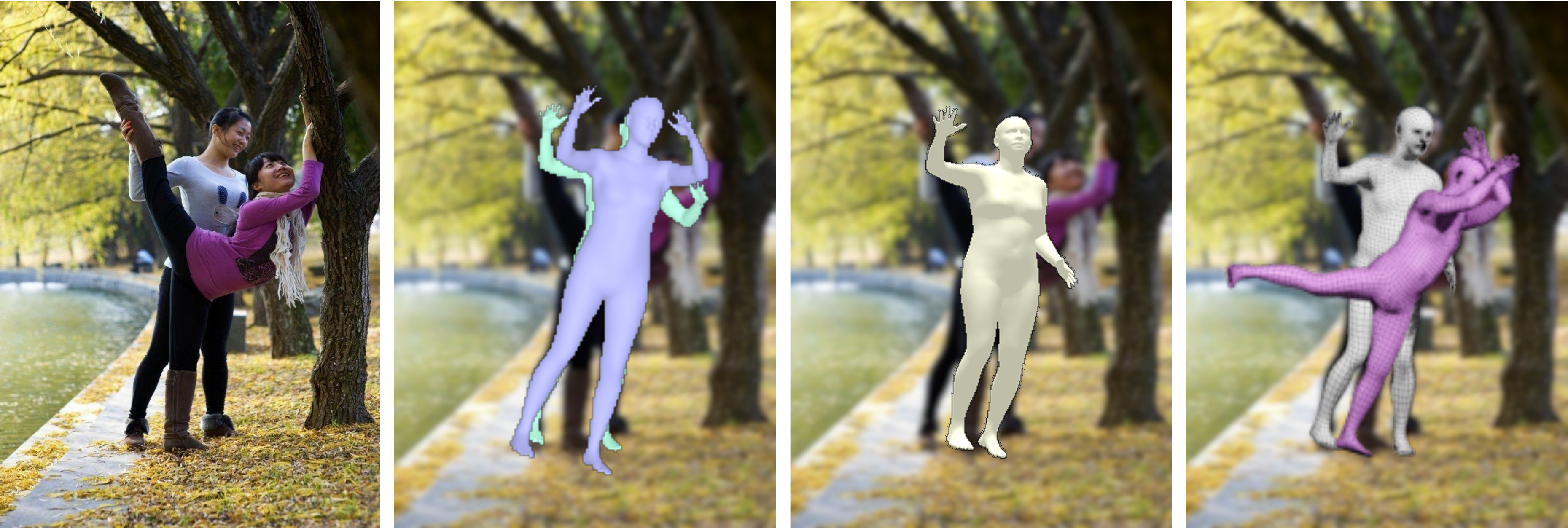}

\includegraphics[height=0.16\textheight,width=0.48\linewidth]{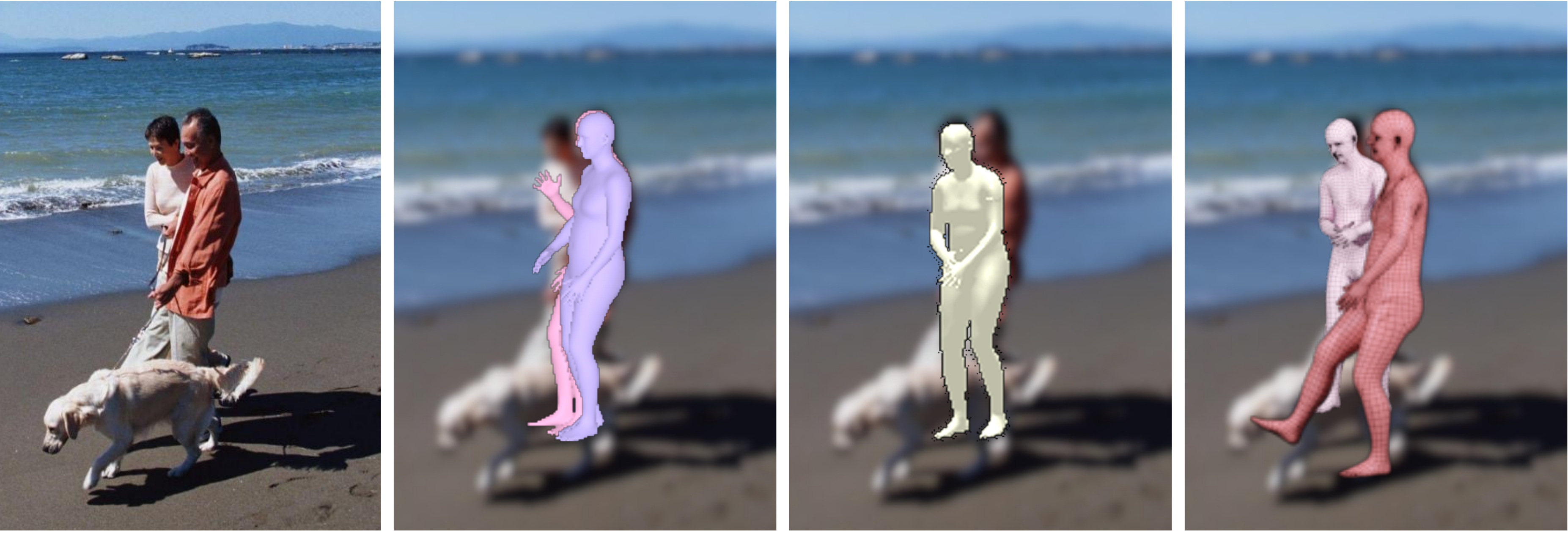}
\includegraphics[height=0.16\textheight,width=0.48\linewidth]{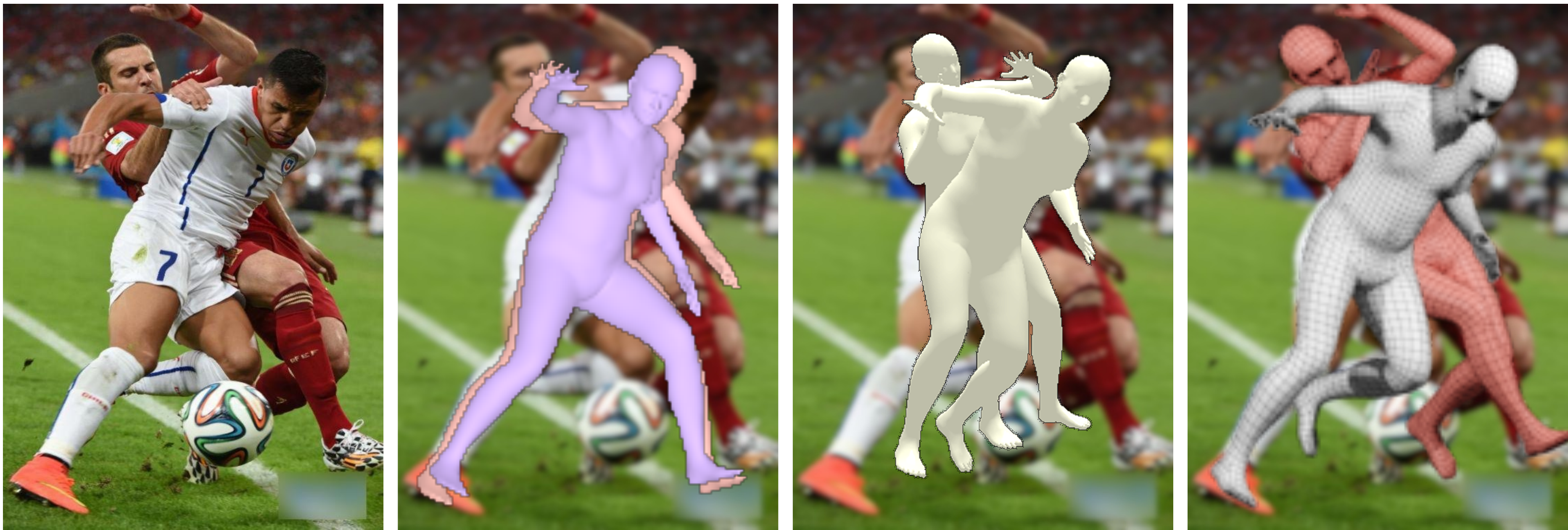}

\includegraphics[height=0.16\textheight,width=0.48\linewidth]{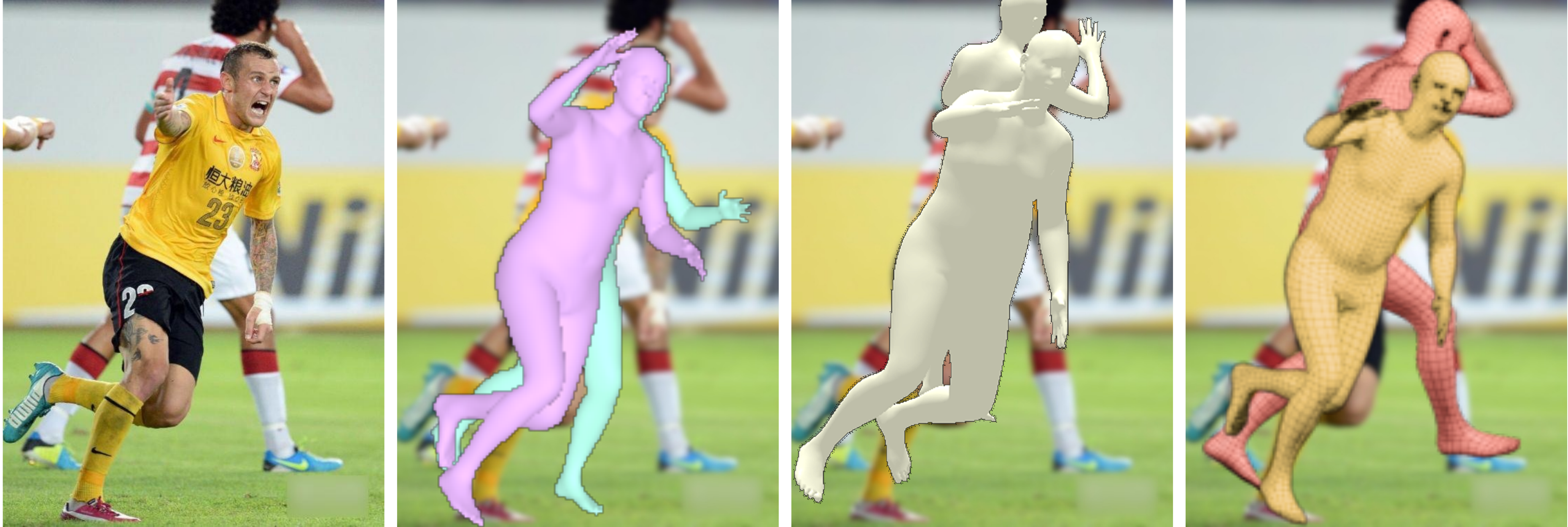}
\includegraphics[height=0.16\textheight,width=0.48\linewidth]{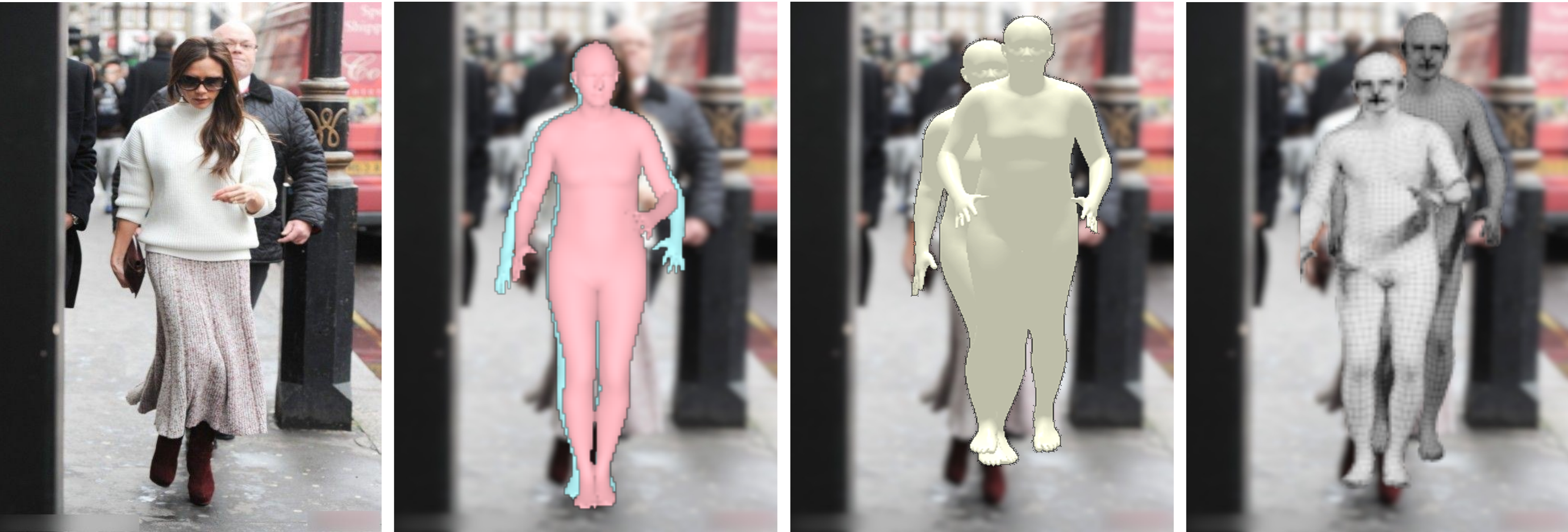}

\includegraphics[height=0.16\textheight,width=0.48\linewidth]{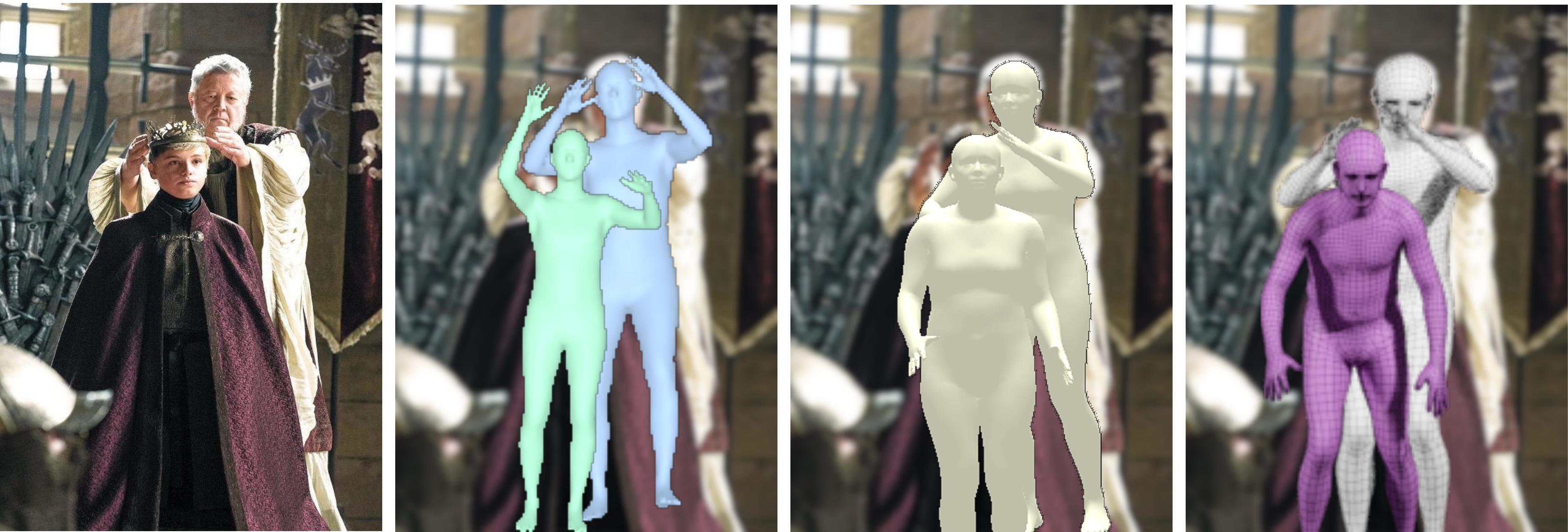}
\includegraphics[height=0.16\textheight,width=0.48\linewidth]{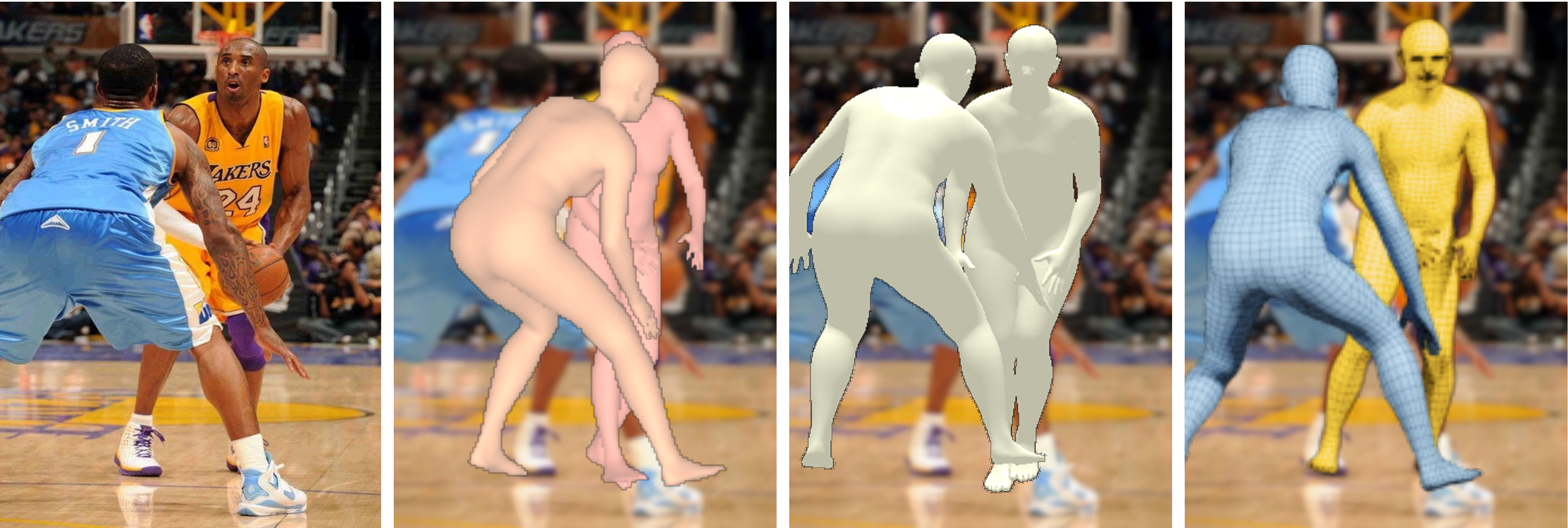}

\includegraphics[height=0.16\textheight,width=0.48\linewidth]{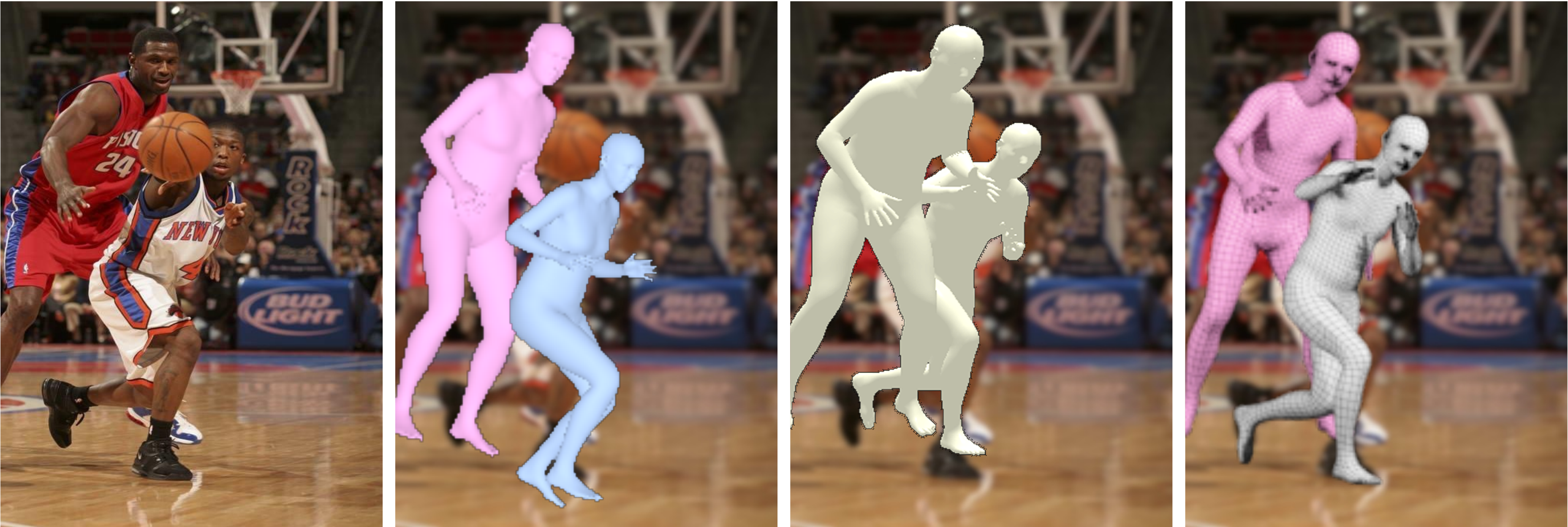}
\includegraphics[height=0.16\textheight,width=0.48\linewidth]{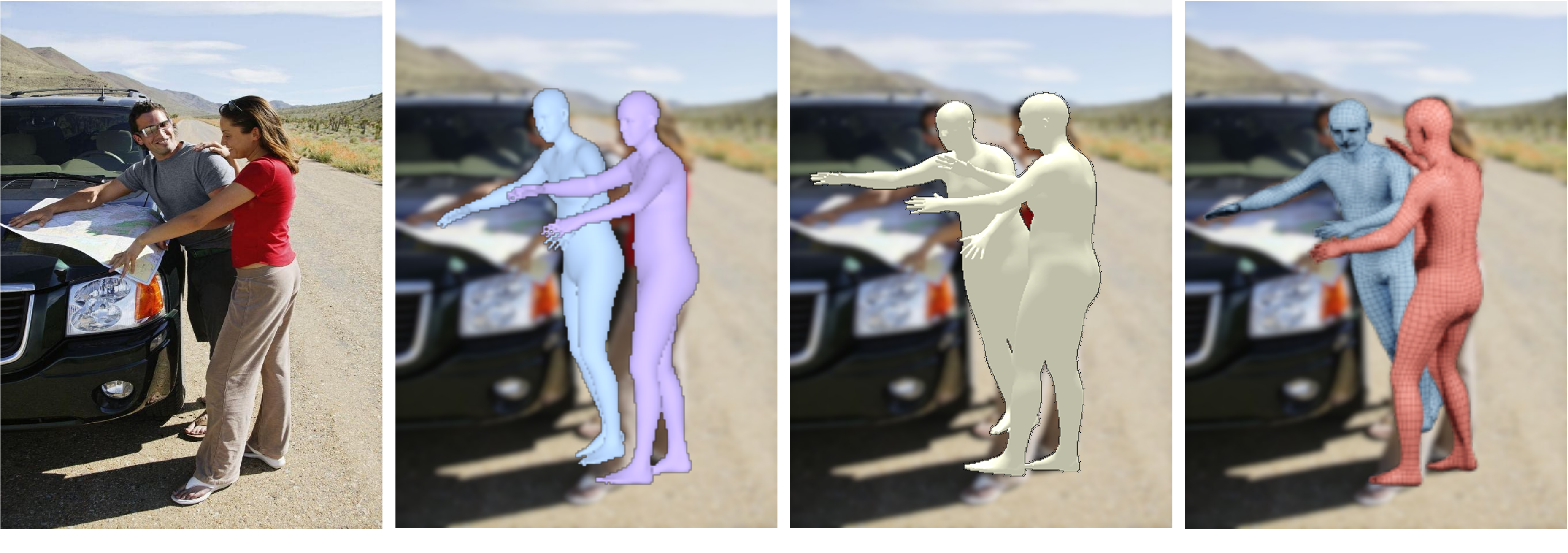}

\end{center}
 \vspace*{-0.2in}
\caption{Qualitative results on the OCHuman \texttt{val} set. Each image (left to right) shows RGB image, SPIN~\cite{kolotouros2019learning} predictions, ROMP~\cite{sun2021monocular} predictions and OCHMR predictions. Due to occlusions, SPIN often misses the person in the background which is recovered by OCHMR. In comparison to ROMP, OCHMR outputs pixel aligned meshes with correct depth-ordering. Please see additional results in Sup. Mat.}
\label{figure:qualitative}
 \end{figure*}

\textbf{Occlusion benchmarks.} To validate the stability under occlusion, we evaluate OCHMR on multiple occlusion
benchmarks. Firstly, on the person-occluded 3DPW-PC, OCHuman and Crowdpose, results in \cref{table:occlusion} show that OCHMR significantly outperforms previous state-of-the-art methods~\cite{sun2021monocular,kolotouros2019learning,zhang2021pymaf}. Additionally, in \cref{figure:qualitative}, we qualitatively demonstrate the robustness of OCHMR under severe occlusion in comparison to top-down SPIN~\cite{kolotouros2019learning} and bottom-up ROMP~\cite{sun2021monocular}. Further, when using ground-truth bounding boxes, the gains of OCHMR are significant in comparison to baselines. These results show that using high-resolution input images along with global/local centermaps is key for occlusion reasoning.

\textbf{General benchmarks.} We also compare OCHMR with other approaches on general benchmarks like 3DPW (\cref{table:3dpw}) and COCO (\cref{table:coco}). OCHMR undergoes no performance degradation on non-occlusion cases. Infact, OCHMR improves baseline SPIN's MPJPE error by $5$mm on 3DPW. Without using extra supervision, our method achieves comparable performance to ROMP with ResNet-50 backbone. We also outperform other methods on the COCO dataset.

\begin{table}[b]
    \vspace*{-0.2in}
    \centering
    \small
    \resizebox{3.4in}{!}{
        \setlength\tabcolsep{4pt}
        \renewcommand{\arraystretch}{1.2} 
        \rowcolors{1}{}{lightgray}
        \begin{tabular}{@{}l|c|l l l}
        \Xhline{3\arrayrulewidth}
        \textbf{Method} & \textbf{H3.6M} &\textbf{MPJPE$\downarrow$} & \textbf{PMPJPE$\downarrow$} &  \textbf{PVE$\downarrow$}\\
        \hline 
        HMR~\cite{kanazawa2018end} & \cmark & 130.0 & 76.7 & - \\
        Kanazawa et al.~\cite{kanazawa2019learning} & \cmark & 116.5 & 72.6 & 139.3 \\
        Arnab et al.~\cite{arnab2019exploiting} & \cmark & - & 72.2 & - \\
        GCMR~\cite{kolotouros2019convolutional} & \cmark & - & 70.2 & - \\
        DSD-SATN~\cite{sun2019human} & \cmark & - & 69.5 & -\\
        SPIN~\cite{kolotouros2019learning} & \cmark & 96.9 & 59.2 & 116.4\\
        ROMP (ResNet-50)~\cite{sun2021monocular} & \cmark & 91.3 & 54.9 & 108.3\\
        \hline
        I2L-MeshNet*~\cite{moon2020i2l} & \cmark & 93.2 & 58.6 & -\\
        EFT*~\cite{joo2020exemplar} & \cmark & - & 54.2 & -\\
        VIBE*~\cite{kocabas2020vibe} & \cmark & 93.5 & 56.5 & 113.4\\
        PyMaf*~\cite{zhang2021pymaf} & \cmark & 92.8 & 58.9 & 110.1 \\
        ROMP (ResNet-50)*~\cite{sun2021monocular} & \cmark & \textbf{89.3} & \textbf{53.5} & \textbf{105.6} \\
        \hline

        SPIN~\cite{kolotouros2019learning} & \xmark & 94.7 & 60.2 & 111.4 \\
        OCHMR (Ours) & \xmark & \textbf{89.7 (-5.0)}  & \textbf{58.3 (-1.9)} & \textbf{107.1 (-4.3)}\\
            \Xhline{3\arrayrulewidth}
        \end{tabular}
    }

    \caption{Comparisons to the state-of-the-art methods on 3DPW \texttt{test} set using \textit{Protocol 2}~\cite{sun2021monocular}. * denotes extra training data in comparison to SPIN~\cite{kolotouros2019learning}. OCHMR does not use Human3.6M~\cite{ionescu2013human3} for training and achieves comparable results to prior art that uses extra supervision. }
    \label{table:3dpw}
\end{table}

\begin{table}
    \small
    \centering
    \resizebox{3.1in}{!}{
    \renewcommand{\arraystretch}{1.2} 
    \setlength\tabcolsep{3pt}
    \rowcolors{1}{}{lightgray}
    \begin{tabular}{@{}l|l|c c c c c@{}}
\Xhline{3\arrayrulewidth}
\textbf{Method}  & \textbf{$\text{AP}$}$\uparrow$ & \textbf{$\text{AP}^{50}$}$\uparrow$ & \textbf{$\text{AP}^{75}$}$\uparrow$ & \textbf{$\text{AP}^\text{M}$}$\uparrow$ & \textbf{$\text{AP}^\text{L}$}$\uparrow$ &\textbf{$\text{AR}$}$\uparrow$  \\
    \hline
    SPIN~\cite{kolotouros2019learning}  &   11.3 & 28.6 & 5.8 & 10.2 & 11.4 & 22.8\\
     CRMH$\star$~\cite{jiang2020coherent}  &  12.6 & 33.8 & 7.6 & 13.2 & 12.8 & 25.0\\
      PyMaf$\star$~\cite{zhang2021pymaf}  &  13.8 & 35.8 & 9.7 & 14.8 & 14.2 & 28.9\\
  ROMP~\cite{sun2021monocular} & 14.7 & 36.7 & 9.8 & 15.3 & 14.8 & 29.0\\
    OCHMR (ours) & \textbf{15.3 (+0.6)} & \textbf{38.7} & \textbf{10.2} & \textbf{16.7} & \textbf{15.9} & \textbf{29.4}\\
    \hline
    \multicolumn{7}{c}{Using ground-truth bounding boxes} \\
    \hline
      SPIN~\cite{kolotouros2019learning}  & 
      13.0 & 33.8 & 7.0 & 13.6 & 12.9 & 26.8\\
 OCHMR (Ours) & 
      \textbf{17.4 (+4.4)} & \textbf{41.9} & \textbf{11.8} & \textbf{18.2} & \textbf{17.4} & \textbf{32.4}\\

\Xhline{3\arrayrulewidth}
  \end{tabular}
   }
    \caption{Comparisons to the state-of-the-art methods on COCO \texttt{val} set evaluated for 2D keypoint projection. $\star$ denotes extra training data compared to OCHMR.}
    \label{table:coco}
    \vspace*{-0.1in}

\end{table}

\subsection{Analysis}
We perform all our analysis on the 3DPW-PC dataset with ground-truth boxes for evaluations.

\noindent
\textbf{CoNorm Block Architecture.} We compare CoNorm blocks against \textit{early} and \textit{late} fusion in \cref{table:conorm_arch}. In \textit{early} fusion, we perform channel-wise concatenation of input image, global centermap and local centermap. In \textit{late} fusion we concatenate the intermediate feature after the third ResNet block with the downsampled context information. We observe that injection of high-resolution context information at multiple-depths in the form of CoNorm blocks is important for accurate human mesh recovery under occlusion. Further, we vary the dimension $K$ of the latent space of the four CoNorm blocks in the  OCHMR backbone. We show that increasing $K$ improves performance under occlusion in comparison to baseline SPIN, with $K=128$ achieving the optimal balance between the parameter overhead and human recovery performance.

\begin{table}[b]
\centering
\small
\resizebox{3.0in}{!}{
    \renewcommand{\arraystretch}{1.2}
    \rowcolors{1}{}{lightgray}
    \setlength{\tabcolsep}{6pt}
    \begin{tabular}{@{}l|c c c @{}}
        \Xhline{3\arrayrulewidth}
        \textbf{Method} & \textbf{MPJPE $\downarrow$} & \textbf{PMPJPE$\downarrow$} & \textbf{PVE$\downarrow$} \\
        \hline
        SPIN & 128.4 & 82.1 & 155.7  \\
        OCHMR, \textit{early-fusion} & 115.8 & 76.4 & 150.1  \\ 
        OCHMR, \textit{late-fusion} & 119.8 & 80.2 & 151.8  \\
        \hline
        OCHMR, $K=16$ & 116.8 & 76.9 & 150.2 \\
        OCHMR, $K=32$ & 114.2 & 76.2 & 148.6 \\
        OCHMR, $K=64$ & 113.0 & 75.0 & 146.4 \\
        OCHMR, $K=128$ & \textbf{112.2} & 75.2 & \textbf{145.9} \\
        OCHMR, $K=256$ & 113.1 & \textbf{74.7} & 146.1\\
        \Xhline{3\arrayrulewidth}
    \end{tabular}
}
\caption{Comparison of CoNorm block with \textit{early} and \textit{late} fusion of context along with variation of CoNorm block's latent space dimension $K$. Injection of contextual information at multiple depths outperforms \textit{early}/\textit{late} fusion. Increase in $K$ results in better context normalization with better performance under occlusion.}
\label{table:conorm_arch}
\end{table}



\noindent
\textbf{Effect of Multi-Person losses.} To understand the effect of multi-person losses like interpenetration loss $\mathcal{L}_{\text{collision}}$ and depth-ordering loss $\mathcal{L}_{\text{depth}}$, we perform an ablative study using the loss weights in the OCHMR framework in \cref{table:losses}. We achieve the best performance when using both losses, however the use of supervised $\mathcal{L}_{\text{depth}}$ loss gives better gains than self-supervised $\mathcal{L}_{\text{collision}}$. Note, OCHMR still significantly outperforms baseline SPIN when only using $\mathcal{L}_{\text{single}}$.

\begin{table}[h]
\centering
\small
\resizebox{2.8in}{!}{
    \renewcommand{\arraystretch}{1.2}
    \rowcolors{1}{}{lightgray}
    \setlength{\tabcolsep}{4pt}
  
    \begin{tabular}{@{}c|c|c|c c c @{}}
        \Xhline{3\arrayrulewidth}
       $\mathcal{L}_\text{single}$ & $\mathcal{L}_\text{collision}$ & $\mathcal{L}_\text{depth}$ & \textbf{MPJPE $\downarrow$} & \textbf{PMPJPE$\downarrow$} & \textbf{PVE$\downarrow$} \\
        \hline
        \cmark &\xmark &\xmark & 116.9 & 77.1 & 149.2  \\ 
        \cmark &\cmark &\xmark & 115.3 & 76.2 & 148.7  \\ 
        \cmark &\xmark &\cmark & 113.6 & 75.6 & 147.0  \\ 
        \cmark &\cmark &\cmark & \textbf{112.2} & \textbf{75.2} & \textbf{145.9}  \\ 
        \Xhline{3\arrayrulewidth}
    \end{tabular}
}
\caption{Ablation of multi-person losses in OCHMR. We default $w_\text{single}$ to 1 to ensure model convergence. We found the relative importance of $\mathcal{L}_{\text{depth}}$ to be greater than $\mathcal{L}_{\text{collision}}$.}
\label{table:losses}
\vspace*{-0.1in}
\end{table}

\noindent
\textbf{Choice of Context.} CoNorm blocks allow conditioning the network $P$ with various representations of the spatial-context $\mathbf{C}$. \cref{table:choice_of_context} shows the effect of using ground-truth and predicted (using F) Local and Local + Global Centermaps along with 2D keypoints as $\mathbf{C}$. We use offshelf pose-estimation network HRNet-W48~\cite{sun2019deep} trained on COCO dataset as our $F$. In case of 2D keypoints, $\mathbf{C}$ is a $17$-channel heatmap corresponding to keypoint locations. In comparison to local centermaps, the addition of global centermaps helps improve performance under occlusion. Interestingly, conditioning using ground-truth 2D keypoints outperforms all other choices. However, when ground-truth keypoints are unavailable, body centers outperform estimated 2D keypoints as estimating accurate 2D pose under occlusion is more challenging than estimating body centers. 

\begin{table}[h]
\centering
\small\
\resizebox{3.0in}{!}{
    \renewcommand{\arraystretch}{1.2}
    \rowcolors{3}{}{lightgray}
    \setlength{\tabcolsep}{1pt}
    \begin{tabular}{@{}l|c c | c c@{}}
        \Xhline{3\arrayrulewidth}
        \multirow{2}{*}{\textbf{Context $\mathbf{C}$}} & \multicolumn{2}{c|}{\textbf{Ground-Truth}} & \multicolumn{2}{c}{\textbf{Estimated by} $F$}  \\
         & \textbf{MPJPE $\downarrow$} & \textbf{PMPJPE$\downarrow$} & \textbf{MPJPE $\downarrow$} & \textbf{PMPJPE$\downarrow$}\\
        \hline
        Local Center & 113.0 & 76.4 & 114.8 & 77.1 \\
        Local + Global Center & 111.4 & 74.7 & \textbf{112.2} & \textbf{75.2} \\
        2D Keypoints & \textbf{109.5} & \textbf{73.9} & 116.8 & 78.9 \\
        \Xhline{3\arrayrulewidth}
    \end{tabular}
}
\caption{Comparison of various choices of contexts in OCHMR conditioning. Local + Global centermaps performs better than other conditioning choices when being estimated by the network $F$.}
\label{table:choice_of_context}
\end{table}

\noindent
\textbf{Limitations.} OCHMR is a multi-stage top-down method and is, therefore, not real-time during inference. Though OCHMR improves performance under multi-person occlusion, it is still susceptible to failure under truncation and extreme cropping due to object occlusion. Moreover, OCHMR fails to handle extreme poses and shapes due to the lack of training data, as shown in the Sup. Mat. In the future, OCHMR can be extended and incorporated with the recent progress to handle various kinds of occlusions ~\cite{kocabas2021pare, zhang2020object, pavlakos2020human}.

\section{Conclusion}
\label{sec:conclusion}
Most top-down methods for human mesh recovery assume a single subject in the input, causing them to fail under severe person-person occlusion. In this work, we introduce OCHMR, a novel top-down method to handle multiple occluded people in crowded scenes. Our key idea is conditioning top-down models on spatial-context from the image, in the form of local and global centermaps which allows OCHMR to effectively disambiguate between overlapping humans. We propose Contextual Normalization (CoNorm) blocks, a novel architectural improvement which can be easily extended to any existing top-down method. While OCHMR draws inspiration from bottom-up methods, we retain the advantages of both bottom-up and top-down methods, resulting in a method that can handle multi-person occlusion and achieve pixel-aligned reconstruction results.

\label{sec:experiments_qualitative}

{\small
\bibliographystyle{ieee_fullname}
\bibliography{references}

\begin{thebibliography}{10}\itemsep=-1pt

\bibitem{andriluka20142d}
Mykhaylo Andriluka, Leonid Pishchulin, Peter Gehler, and Bernt Schiele.
\newblock 2d human pose estimation: New benchmark and state of the art
  analysis.
\newblock In {\em Proceedings of the IEEE Conference on computer Vision and
  Pattern Recognition}, pages 3686--3693, 2014.

\bibitem{arnab2019exploiting}
Anurag Arnab, Carl Doersch, and Andrew Zisserman.
\newblock Exploiting temporal context for 3d human pose estimation in the wild.
\newblock In {\em Proceedings of the IEEE/CVF Conference on Computer Vision and
  Pattern Recognition}, pages 3395--3404, 2019.

\bibitem{bochkovskiy2020yolov4}
Alexey Bochkovskiy, Chien-Yao Wang, and Hong-Yuan~Mark Liao.
\newblock Yolov4: Optimal speed and accuracy of object detection.
\newblock {\em arXiv preprint arXiv:2004.10934}, 2020.

\bibitem{bogo2016keep}
Federica Bogo, Angjoo Kanazawa, Christoph Lassner, Peter Gehler, Javier Romero,
  and Michael~J Black.
\newblock Keep it smpl: Automatic estimation of 3d human pose and shape from a
  single image.
\newblock In {\em European conference on computer vision}, pages 561--578.
  Springer, 2016.

\bibitem{cheng2018revisiting}
Bowen Cheng, Yunchao Wei, Honghui Shi, Rogerio Feris, Jinjun Xiong, and Thomas
  Huang.
\newblock Revisiting rcnn: On awakening the classification power of faster
  rcnn.
\newblock In {\em Proceedings of the European conference on computer vision
  (ECCV)}, pages 453--468, 2018.

\bibitem{cheng2020higherhrnet}
Bowen Cheng, Bin Xiao, Jingdong Wang, Honghui Shi, Thomas~S Huang, and Lei
  Zhang.
\newblock Higherhrnet: Scale-aware representation learning for bottom-up human
  pose estimation.
\newblock In {\em Proceedings of the IEEE/CVF Conference on Computer Vision and
  Pattern Recognition}, pages 5386--5395, 2020.

\bibitem{choi2020pose2mesh}
Hongsuk Choi, Gyeongsik Moon, and Kyoung~Mu Lee.
\newblock Pose2mesh: Graph convolutional network for 3d human pose and mesh
  recovery from a 2d human pose.
\newblock In {\em European Conference on Computer Vision}, pages 769--787.
  Springer, 2020.

\bibitem{duan2019centernet}
Kaiwen Duan, Song Bai, Lingxi Xie, Honggang Qi, Qingming Huang, and Qi Tian.
\newblock Centernet: Keypoint triplets for object detection.
\newblock In {\em Proceedings of the IEEE/CVF International Conference on
  Computer Vision}, pages 6569--6578, 2019.

\bibitem{dumoulin2016learned}
Vincent Dumoulin, Jonathon Shlens, and Manjunath Kudlur.
\newblock A learned representation for artistic style.
\newblock {\em arXiv preprint arXiv:1610.07629}, 2016.

\bibitem{dwivedi2021learning}
Sai~Kumar Dwivedi, Nikos Athanasiou, Muhammed Kocabas, and Michael~J Black.
\newblock Learning to regress bodies from images using differentiable semantic
  rendering.
\newblock In {\em Proceedings of the IEEE/CVF International Conference on
  Computer Vision}, pages 11250--11259, 2021.

\bibitem{guan2009estimating}
Peng Guan, Alexander Weiss, Alexandru~O Balan, and Michael~J Black.
\newblock Estimating human shape and pose from a single image.
\newblock In {\em 2009 IEEE 12th International Conference on Computer Vision},
  pages 1381--1388. IEEE, 2009.

\bibitem{guan2021bilevel}
Shanyan Guan, Jingwei Xu, Yunbo Wang, Bingbing Ni, and Xiaokang Yang.
\newblock Bilevel online adaptation for out-of-domain human mesh
  reconstruction.
\newblock In {\em Proceedings of the IEEE/CVF Conference on Computer Vision and
  Pattern Recognition}, pages 10472--10481, 2021.

\bibitem{hassan2019resolving}
Mohamed Hassan, Vasileios Choutas, Dimitrios Tzionas, and Michael~J Black.
\newblock Resolving 3d human pose ambiguities with 3d scene constraints.
\newblock In {\em Proceedings of the IEEE/CVF International Conference on
  Computer Vision}, pages 2282--2292, 2019.

\bibitem{he2017mask}
Kaiming He, Georgia Gkioxari, Piotr Doll{\'a}r, and Ross Girshick.
\newblock Mask r-cnn.
\newblock In {\em Proceedings of the IEEE international conference on computer
  vision}, pages 2961--2969, 2017.

\bibitem{he2016deep}
Kaiming He, Xiangyu Zhang, Shaoqing Ren, and Jian Sun.
\newblock Deep residual learning for image recognition.
\newblock In {\em Proceedings of the IEEE conference on computer vision and
  pattern recognition}, pages 770--778, 2016.

\bibitem{huang2017arbitrary}
Xun Huang and Serge Belongie.
\newblock Arbitrary style transfer in real-time with adaptive instance
  normalization.
\newblock In {\em Proceedings of the IEEE International Conference on Computer
  Vision}, pages 1501--1510, 2017.

\bibitem{ioffe2015batch}
Sergey Ioffe and Christian Szegedy.
\newblock Batch normalization: Accelerating deep network training by reducing
  internal covariate shift.
\newblock In {\em International conference on machine learning}, pages
  448--456. PMLR, 2015.

\bibitem{ionescu2013human3}
Catalin Ionescu, Dragos Papava, Vlad Olaru, and Cristian Sminchisescu.
\newblock Human3. 6m: Large scale datasets and predictive methods for 3d human
  sensing in natural environments.
\newblock {\em IEEE transactions on pattern analysis and machine intelligence},
  36(7):1325--1339, 2013.

\bibitem{jiang2020coherent}
Wen Jiang, Nikos Kolotouros, Georgios Pavlakos, Xiaowei Zhou, and Kostas
  Daniilidis.
\newblock Coherent reconstruction of multiple humans from a single image.
\newblock In {\em Proceedings of the IEEE/CVF Conference on Computer Vision and
  Pattern Recognition}, pages 5579--5588, 2020.

\bibitem{johnson2011learning}
Sam Johnson and Mark Everingham.
\newblock Learning effective human pose estimation from inaccurate annotation.
\newblock In {\em CVPR 2011}, pages 1465--1472. IEEE, 2011.

\bibitem{joo2015panoptic}
Hanbyul Joo, Hao Liu, Lei Tan, Lin Gui, Bart Nabbe, Iain Matthews, Takeo
  Kanade, Shohei Nobuhara, and Yaser Sheikh.
\newblock Panoptic studio: A massively multiview system for social motion
  capture.
\newblock In {\em Proceedings of the IEEE International Conference on Computer
  Vision}, pages 3334--3342, 2015.

\bibitem{joo2020exemplar}
Hanbyul Joo, Natalia Neverova, and Andrea Vedaldi.
\newblock Exemplar fine-tuning for 3d human model fitting towards in-the-wild
  3d human pose estimation.
\newblock {\em arXiv preprint arXiv:2004.03686}, 2020.

\bibitem{joo2018total}
Hanbyul Joo, Tomas Simon, and Yaser Sheikh.
\newblock Total capture: A 3d deformation model for tracking faces, hands, and
  bodies.
\newblock In {\em Proceedings of the IEEE conference on computer vision and
  pattern recognition}, pages 8320--8329, 2018.

\bibitem{kanazawa2017end}
Angjoo Kanazawa, Michael~J Black, David~W Jacobs, and Jitendra Malik.
\newblock End-to-end recovery of human shape and pose. corr abs/1712.06584
  (2017).
\newblock {\em arXiv preprint arXiv:1712.06584}, 2017.

\bibitem{kanazawa2018end}
Angjoo Kanazawa, Michael~J Black, David~W Jacobs, and Jitendra Malik.
\newblock End-to-end recovery of human shape and pose.
\newblock In {\em Proceedings of the IEEE conference on computer vision and
  pattern recognition}, pages 7122--7131, 2018.

\bibitem{kanazawa2019learning}
Angjoo Kanazawa, Jason~Y Zhang, Panna Felsen, and Jitendra Malik.
\newblock Learning 3d human dynamics from video.
\newblock In {\em Proceedings of the IEEE/CVF Conference on Computer Vision and
  Pattern Recognition}, pages 5614--5623, 2019.

\bibitem{kato2018neural}
Hiroharu Kato, Yoshitaka Ushiku, and Tatsuya Harada.
\newblock Neural 3d mesh renderer.
\newblock In {\em Proceedings of the IEEE conference on computer vision and
  pattern recognition}, pages 3907--3916, 2018.

\bibitem{khirodkar2021multi}
Rawal Khirodkar, Visesh Chari, Amit Agrawal, and Ambrish Tyagi.
\newblock Multi-hypothesis pose networks: Rethinking top-down pose estimation.
\newblock {\em arXiv preprint arXiv:2101.11223}, 2021.

\bibitem{kissos2020beyond}
Imry Kissos, Lior Fritz, Matan Goldman, Omer Meir, Eduard Oks, and Mark Kliger.
\newblock Beyond weak perspective for monocular 3d human pose estimation.
\newblock In {\em European Conference on Computer Vision}, pages 541--554.
  Springer, 2020.

\bibitem{kocabas2020vibe}
Muhammed Kocabas, Nikos Athanasiou, and Michael~J Black.
\newblock Vibe: Video inference for human body pose and shape estimation.
\newblock In {\em Proceedings of the IEEE/CVF Conference on Computer Vision and
  Pattern Recognition}, pages 5253--5263, 2020.

\bibitem{kocabas2021pare}
Muhammed Kocabas, Chun-Hao~P Huang, Otmar Hilliges, and Michael~J Black.
\newblock Pare: Part attention regressor for 3d human body estimation.
\newblock {\em arXiv preprint arXiv:2104.08527}, 2021.

\bibitem{kolotouros2019learning}
Nikos Kolotouros, Georgios Pavlakos, Michael~J Black, and Kostas Daniilidis.
\newblock Learning to reconstruct 3d human pose and shape via model-fitting in
  the loop.
\newblock In {\em Proceedings of the IEEE/CVF International Conference on
  Computer Vision}, pages 2252--2261, 2019.

\bibitem{kolotouros2019convolutional}
Nikos Kolotouros, Georgios Pavlakos, and Kostas Daniilidis.
\newblock Convolutional mesh regression for single-image human shape
  reconstruction.
\newblock In {\em Proceedings of the IEEE/CVF Conference on Computer Vision and
  Pattern Recognition}, pages 4501--4510, 2019.

\bibitem{li2019crowdpose}
Jiefeng Li, Can Wang, Hao Zhu, Yihuan Mao, Hao-Shu Fang, and Cewu Lu.
\newblock Crowdpose: Efficient crowded scenes pose estimation and a new
  benchmark.
\newblock In {\em Proceedings of the IEEE/CVF Conference on Computer Vision and
  Pattern Recognition}, pages 10863--10872, 2019.

\bibitem{li2021hybrik}
Jiefeng Li, Chao Xu, Zhicun Chen, Siyuan Bian, Lixin Yang, and Cewu Lu.
\newblock Hybrik: A hybrid analytical-neural inverse kinematics solution for 3d
  human pose and shape estimation.
\newblock In {\em Proceedings of the IEEE/CVF Conference on Computer Vision and
  Pattern Recognition}, pages 3383--3393, 2021.

\bibitem{lin2021end}
Kevin Lin, Lijuan Wang, and Zicheng Liu.
\newblock End-to-end human pose and mesh reconstruction with transformers.
\newblock In {\em Proceedings of the IEEE/CVF Conference on Computer Vision and
  Pattern Recognition}, pages 1954--1963, 2021.

\bibitem{lin2021-mesh-graphormer}
Kevin Lin, Lijuan Wang, and Zicheng Liu.
\newblock Mesh graphormer.
\newblock In {\em ICCV}, 2021.

\bibitem{lin2017feature}
Tsung-Yi Lin, Piotr Doll{\'a}r, Ross Girshick, Kaiming He, Bharath Hariharan,
  and Serge Belongie.
\newblock Feature pyramid networks for object detection.
\newblock In {\em Proceedings of the IEEE conference on computer vision and
  pattern recognition}, pages 2117--2125, 2017.

\bibitem{lin2014microsoft}
Tsung-Yi Lin, Michael Maire, Serge Belongie, James Hays, Pietro Perona, Deva
  Ramanan, Piotr Doll{\'a}r, and C~Lawrence Zitnick.
\newblock Microsoft coco: Common objects in context.
\newblock In {\em European conference on computer vision}, pages 740--755.
  Springer, 2014.

\bibitem{liu2021polarized}
Huajun Liu, Fuqiang Liu, Xinyi Fan, and Dong Huang.
\newblock Polarized self-attention: Towards high-quality pixel-wise regression.
\newblock {\em arXiv preprint arXiv:2107.00782}, 2021.

\bibitem{liu2016ssd}
Wei Liu, Dragomir Anguelov, Dumitru Erhan, Christian Szegedy, Scott Reed,
  Cheng-Yang Fu, and Alexander~C Berg.
\newblock Ssd: Single shot multibox detector.
\newblock In {\em European conference on computer vision}, pages 21--37.
  Springer, 2016.

\bibitem{long2015fully}
Jonathan Long, Evan Shelhamer, and Trevor Darrell.
\newblock Fully convolutional networks for semantic segmentation.
\newblock In {\em Proceedings of the IEEE conference on computer vision and
  pattern recognition}, pages 3431--3440, 2015.

\bibitem{loper2015smpl}
Matthew Loper, Naureen Mahmood, Javier Romero, Gerard Pons-Moll, and Michael~J
  Black.
\newblock Smpl: A skinned multi-person linear model.
\newblock {\em ACM transactions on graphics (TOG)}, 34(6):1--16, 2015.

\bibitem{ma2021towards}
Zhiheng Ma, Xiaopeng Hong, Xing Wei, Yunfeng Qiu, and Yihong Gong.
\newblock Towards a universal model for cross-dataset crowd counting.
\newblock In {\em Proceedings of the IEEE/CVF International Conference on
  Computer Vision}, pages 3205--3214, 2021.

\bibitem{mcnally2021evopose2d}
William McNally, Kanav Vats, Alexander Wong, and John McPhee.
\newblock Evopose2d: Pushing the boundaries of 2d human pose estimation using
  accelerated neuroevolution with weight transfer.
\newblock {\em IEEE Access}, 2021.

\bibitem{mehta2017monocular}
Dushyant Mehta, Helge Rhodin, Dan Casas, Pascal Fua, Oleksandr Sotnychenko,
  Weipeng Xu, and Christian Theobalt.
\newblock Monocular 3d human pose estimation in the wild using improved cnn
  supervision.
\newblock In {\em 2017 international conference on 3D vision (3DV)}, pages
  506--516. IEEE, 2017.

\bibitem{moon2020i2l}
Gyeongsik Moon and Kyoung~Mu Lee.
\newblock I2l-meshnet: Image-to-lixel prediction network for accurate 3d human
  pose and mesh estimation from a single rgb image.
\newblock In {\em Computer Vision--ECCV 2020: 16th European Conference,
  Glasgow, UK, August 23--28, 2020, Proceedings, Part VII 16}, pages 752--768.
  Springer, 2020.

\bibitem{omran2018neural}
Mohamed Omran, Christoph Lassner, Gerard Pons-Moll, Peter Gehler, and Bernt
  Schiele.
\newblock Neural body fitting: Unifying deep learning and model based human
  pose and shape estimation.
\newblock In {\em 2018 international conference on 3D vision (3DV)}, pages
  484--494. IEEE, 2018.

\bibitem{pavlakos2020human}
Georgios Pavlakos, Jitendra Malik, and Angjoo Kanazawa.
\newblock Human mesh recovery from multiple shots.
\newblock {\em arXiv preprint arXiv:2012.09843}, 2020.

\bibitem{pavlakos2018learning}
Georgios Pavlakos, Luyang Zhu, Xiaowei Zhou, and Kostas Daniilidis.
\newblock Learning to estimate 3d human pose and shape from a single color
  image.
\newblock In {\em Proceedings of the IEEE Conference on Computer Vision and
  Pattern Recognition}, pages 459--468, 2018.

\bibitem{perez2018film}
Ethan Perez, Florian Strub, Harm De~Vries, Vincent Dumoulin, and Aaron
  Courville.
\newblock Film: Visual reasoning with a general conditioning layer.
\newblock In {\em Proceedings of the AAAI Conference on Artificial
  Intelligence}, volume~32, 2018.

\bibitem{ren2015faster}
Shaoqing Ren, Kaiming He, Ross Girshick, and Jian Sun.
\newblock Faster r-cnn: Towards real-time object detection with region proposal
  networks.
\newblock {\em Advances in neural information processing systems}, 28:91--99,
  2015.

\bibitem{ruggero2017benchmarking}
Matteo Ruggero~Ronchi and Pietro Perona.
\newblock Benchmarking and error diagnosis in multi-instance pose estimation.
\newblock In {\em Proceedings of the IEEE international conference on computer
  vision}, pages 369--378, 2017.

\bibitem{sengupta2020synthetic}
Akash Sengupta, Ignas Budvytis, and Roberto Cipolla.
\newblock Synthetic training for accurate 3d human pose and shape estimation in
  the wild.
\newblock {\em arXiv preprint arXiv:2009.10013}, 2020.

\bibitem{sigal2010humaneva}
Leonid Sigal, Alexandru~O Balan, and Michael~J Black.
\newblock Humaneva: Synchronized video and motion capture dataset and baseline
  algorithm for evaluation of articulated human motion.
\newblock {\em International journal of computer vision}, 87(1-2):4, 2010.

\bibitem{song2021rethinking}
Qingyu Song, Changan Wang, Zhengkai Jiang, Yabiao Wang, Ying Tai, Chengjie
  Wang, Jilin Li, Feiyue Huang, and Yang Wu.
\newblock Rethinking counting and localization in crowds: A purely point-based
  framework.
\newblock In {\em Proceedings of the IEEE/CVF International Conference on
  Computer Vision}, pages 3365--3374, 2021.

\bibitem{sun2019deep}
Ke Sun, Bin Xiao, Dong Liu, and Jingdong Wang.
\newblock Deep high-resolution representation learning for human pose
  estimation.
\newblock In {\em Proceedings of the IEEE/CVF Conference on Computer Vision and
  Pattern Recognition}, pages 5693--5703, 2019.

\bibitem{sun2021monocular}
Yu Sun, Qian Bao, Wu Liu, Yili Fu, Michael~J Black, and Tao Mei.
\newblock Monocular, one-stage, regression of multiple 3d people.
\newblock In {\em Proceedings of the IEEE/CVF International Conference on
  Computer Vision}, pages 11179--11188, 2021.

\bibitem{sun2019human}
Yu Sun, Yun Ye, Wu Liu, Wenpeng Gao, Yili Fu, and Tao Mei.
\newblock Human mesh recovery from monocular images via a skeleton-disentangled
  representation.
\newblock In {\em Proceedings of the IEEE/CVF International Conference on
  Computer Vision}, pages 5349--5358, 2019.

\bibitem{tripathi2020posenet3d}
Shashank Tripathi, Siddhant Ranade, Ambrish Tyagi, and Amit Agrawal.
\newblock Posenet3d: Learning temporally consistent 3d human pose via knowledge
  distillation.
\newblock In {\em 2020 International Conference on 3D Vision (3DV)}, pages
  311--321. IEEE, 2020.

\bibitem{von2018recovering}
Timo von Marcard, Roberto Henschel, Michael~J Black, Bodo Rosenhahn, and Gerard
  Pons-Moll.
\newblock Recovering accurate 3d human pose in the wild using imus and a moving
  camera.
\newblock In {\em Proceedings of the European Conference on Computer Vision
  (ECCV)}, pages 601--617, 2018.

\bibitem{wang2021uniformity}
Changan Wang, Qingyu Song, Boshen Zhang, Yabiao Wang, Ying Tai, Xuyi Hu,
  Chengjie Wang, Jilin Li, Jiayi Ma, and Yang Wu.
\newblock Uniformity in heterogeneity: Diving deep into count interval
  partition for crowd counting.
\newblock In {\em Proceedings of the IEEE/CVF International Conference on
  Computer Vision}, pages 3234--3242, 2021.

\bibitem{yi2022mover}
Hongwei Yi, Chun-Hao~P. Huang, Dimitrios Tzionas, Muhammed Kocabas, Mohamed
  Hassan, Siyu Tang, Justus Thies, and Michael~J. Black.
\newblock Human-aware object placement for visual environment reconstruction.
\newblock In {\em Computer Vision and Pattern Recognition (CVPR)}, June 2022.

\bibitem{zanfir2018deep}
Andrei Zanfir, Elisabeta Marinoiu, Mihai Zanfir, Alin-Ionut Popa, and Cristian
  Sminchisescu.
\newblock Deep network for the integrated 3d sensing of multiple people in
  natural images.
\newblock {\em Advances in Neural Information Processing Systems},
  31:8410--8419, 2018.

\bibitem{zhang2020distribution}
Feng Zhang, Xiatian Zhu, Hanbin Dai, Mao Ye, and Ce Zhu.
\newblock Distribution-aware coordinate representation for human pose
  estimation.
\newblock In {\em Proceedings of the IEEE/CVF conference on computer vision and
  pattern recognition}, pages 7093--7102, 2020.

\bibitem{zhang2019danet}
Hongwen Zhang, Jie Cao, Guo Lu, Wanli Ouyang, and Zhenan Sun.
\newblock Danet: Decompose-and-aggregate network for 3d human shape and pose
  estimation.
\newblock In {\em Proceedings of the 27th ACM International Conference on
  Multimedia}, pages 935--944, 2019.

\bibitem{zhang2021pymaf}
Hongwen Zhang, Yating Tian, Xinchi Zhou, Wanli Ouyang, Yebin Liu, Limin Wang,
  and Zhenan Sun.
\newblock Pymaf: 3d human pose and shape regression with pyramidal mesh
  alignment feedback loop.
\newblock In {\em Proceedings of the IEEE/CVF International Conference on
  Computer Vision}, pages 11446--11456, 2021.

\bibitem{zhang2021body}
Jianfeng Zhang, Dongdong Yu, Jun~Hao Liew, Xuecheng Nie, and Jiashi Feng.
\newblock Body meshes as points.
\newblock In {\em Proceedings of the IEEE/CVF Conference on Computer Vision and
  Pattern Recognition}, pages 546--556, 2021.

\bibitem{zhang2019pose2seg}
Song-Hai Zhang, Ruilong Li, Xin Dong, Paul Rosin, Zixi Cai, Xi Han, Dingcheng
  Yang, Haozhi Huang, and Shi-Min Hu.
\newblock Pose2seg: Detection free human instance segmentation.
\newblock In {\em Proceedings of the IEEE/CVF Conference on Computer Vision and
  Pattern Recognition}, pages 889--898, 2019.

\bibitem{zhang2020object}
Tianshu Zhang, Buzhen Huang, and Yangang Wang.
\newblock Object-occluded human shape and pose estimation from a single color
  image.
\newblock In {\em Proceedings of the IEEE/CVF Conference on Computer Vision and
  Pattern Recognition}, pages 7376--7385, 2020.

\bibitem{zhou2019objects}
Xingyi Zhou, Dequan Wang, and Philipp Kr{\"a}henb{\"u}hl.
\newblock Objects as points.
\newblock {\em arXiv preprint arXiv:1904.07850}, 2019.

\bibitem{zhou2019bottom}
Xingyi Zhou, Jiacheng Zhuo, and Philipp Krahenbuhl.
\newblock Bottom-up object detection by grouping extreme and center points.
\newblock In {\em Proceedings of the IEEE/CVF Conference on Computer Vision and
  Pattern Recognition}, pages 850--859, 2019.

\bibitem{zhou2019continuity}
Yi Zhou, Connelly Barnes, Jingwan Lu, Jimei Yang, and Hao Li.
\newblock On the continuity of rotation representations in neural networks.
\newblock In {\em Proceedings of the IEEE/CVF Conference on Computer Vision and
  Pattern Recognition}, pages 5745--5753, 2019.

\end{thebibliography}
}

\end{document}


\title{Occluded Human Mesh Recovery}

\author{{Rawal Khirodkar}{\tt\small rkhirodk@cs.cmu.edu}}
\maketitle

\section{Context Normalization (CoNorm) Block Code}
In this section, we describe the code of CoNorm Block in \texttt{PyTorch}. The code in Listing.~\ref{code:conorm} outlines the details of functions $\mathbf{\Phi}_{\text{latent}}$, $\mathbf{\Phi}_{\text{scale}}$ and $\mathbf{\Phi}_{\text{bias}}$. 

\begin{lstlisting}[language=Python, caption=Code for CoNorm block., label=code:conorm]
class CoNorm(nn.Module):
  
    def __init__(self, num_channels, hidden_channels=128):
        """
            CoNorm Block for Occluded Human Mesh Recovery
            num_channels: number of channels in the intermediate feature X
            hidden_channels: K, dimensionality of the latent space of the CoNorm block
        """
        super(CoNorm, self).__init__()
        
        self.phi_latent = nn.Sequential(
                nn.Conv2d(2, hidden_channels, kernel_size=3, padding=1),
                nn.ReLU()
            )
        self.phi_scale = nn.Conv2d(hidden_channels, num_channels, kernel_size=3, padding=1)
        self.phi_bias = nn.Conv2d(hidden_channels, num_channels, kernel_size=3, padding=1)
        return

    def forward(self, X, context):
        """
            X: intemediate feature of the segmentation backbone
            context: local and global centermap concatenated channelwise
        """
        context = F.interpolate(context, size=X.size()[2:], mode='bilinear', align_corners=False)
        
        lambda_ = self.phi_latent(context)
        scale = self.phi_scale(lambda_)
        bias = self.phi_bias(lambda_)
        
        X_prime = X * scale + bias
        
        return X_prime
\end{lstlisting}

\section{Implementation Details}
We use ResNet-50~\cite{he2016deep} backbone for our implementation. For training speedup, we use distributed training using PyTorch~\cite{paszke2019pytorch} using $8$ GeForce RTX 2080 GPUs. The learning rate is set to $5e-5$ and we train for $100$ epochs using SGD optimizer. Similar to ~\cite{kolotouros2019learning}, we resize the input image to $224 \times 224$ and use a batch size of $64$. While training, we use data augmentations like rotation, scaling, brightness and contrast changes. Further, we do not use SMPLify~\cite{loper2015smpl}, static-fits or any pseudo ground-truth when training. The $2D$ and $3D$ loss weights are set similar to ~\cite{kolotouros2019learning}. The context estimator $F$ follows the HRNet-W32~\cite{sun2019deep} backbone and is trained separately on the COCO~\cite{lin2014microsoft} dataset. All the hyper-parameters are set similar to the official implementation of HRNet~\cite{sun2019deep}. We use off-shelf bounding box detector in the form of FasterRCNN~\cite{ren2015faster} from the \textit{detectron2}\footnote{\url{https://github.com/facebookresearch/detectron2}} codebase.


\section{CoNorm Block Placement} 
We analyse the effect of adding CoNorm blocks at various depths in the OCHMR backbone in \cref{table:supplementary:conorm_placement}. Specifically, we investigate insertion after each of the four ResNet blocks for context normalization denoted by : Depth 1\textit{(early)},  Depth 2, Depth 3, Depth 4(\textit{late}). Empirically, inserting after at Depth 2 achieved the best performance with MPJPE of $114.0$mm an improvement of $14.4$mm over baseline. This is helpful when designing parameter efficient conditioning models where injection of information at a particular depth is imporant. This flexibility is provided by the CoNorm blocks.

\begin{table}[h]
\centering
\small
    \renewcommand{\arraystretch}{1.2}
    \rowcolors{1}{}{lightgray}
    \setlength{\tabcolsep}{4pt}
    \begin{tabular}{@{}l|c c c c c@{}}
        \Xhline{3\arrayrulewidth}
         & SPIN & Depth 1  & Depth 2  & Depth 3 & Depth 4\\
        MPJPE $\downarrow$ & 128.4  & 114.8 & \textbf{114.0} & 115.6 & 118.4 \\
        PMPJPE $\downarrow$ &  82.1 & 76.7 & \textbf{76.1} & 78.9 & 80.0 \\
        \Xhline{3\arrayrulewidth}
    \end{tabular}
\caption{Comparison of insertion of a single CoNorm block at various depths in the backbone. Context based feature normalization achieved best performance after the second ResNet block.}
\label{table:supplementary:conorm_placement}
\end{table}


\section{Robustness to Human Detector Outputs} 
The performance of top-down methods is often dependent on the quality of the detected bounding boxes. We analyse the robustness of SPIN and OCHMR with varying detector confidence in \cref{tab:supplementary:detector_conf} on the COCO~\cite{lin2014microsoft}, CrowdPose~\cite{li2019crowdpose}, OCHuman~\cite{zhang2019pose2seg} datasets using FasterRCNN~\cite{ren2015faster}.

\vspace{0.5cm}

OCHMR outperforms SPIN at all confidence thresholds, especially at stricter confidence thresholds on all datasets. This is consistent with our gains shown with ground-truth bounding boxes in the main paper.


\begin{table}[H]
    \centering
    \renewcommand{\arraystretch}{1.0} 
    \begin{tabular}{@{}c | r c c | r c c | r c c@{}}
    \hline
    \textbf{Min. BB}   & \multicolumn{3}{c|}{\textbf{COCO}}  & \multicolumn{3}{c|}{\textbf{CrowdPose}} & \multicolumn{3}{c}{\textbf{OCHuman}} \\
    Confid. & \#boxes & SPIN(AP) & OCHMR(AP) & \#boxes & SPIN(AP) & OCHMR(AP) & \#boxes & SPIN(AP) & OCHMR(AP)\\
    \hline
    0.0     &   104125  & 11.3  &  15.3 &  27480   & 14.8 & 21.4  &   30637   & 11.2 & 24.8    \\
    0.1     &   44346   & 10.7 & 14.6 & 21237   & 13.4 & 20.1  &   22247   & 10.6  &  23.5 \\
    0.2     &   31748   & 10.4 & 14.3 & 16594   & 13.0 & 19.5    &   16273   & 10.4    & 22.8 \\
    0.3     &   24848   & 9.4 & 14.1  & 14358   &  12.5  & 19.0 &   13603   &  9.5 &  21.8 \\
    0.4     &   17534   & 8.7 & 13.6 & 12968   & 12.1  & 18.6 &   11944   & 9.2  &  21.5 \\
    0.5     &   14381   & 7.9 & 13.2 & 11869   & 11.5 &  18.5 &   10654   & 8.6  &  21.4 \\
    0.6     &   12794   & 7.5 & 12.6 & 10868   & 11.2 &  18.0  &   9626   &  8.1  & 21.3 \\
    0.7     &   11712   & 7.1 & 12.4 & 9944    & 11.2    & 17.2 &   8699   &  7.6  & 21.3 \\
    0.8     &   10631   & 6.7 & 12.4 & 8964    & 10.6  & 16.4  &   7768   &  7.3  & 20.8 \\
    0.9     &   9053    & 6.5 & 12.3 & 7780    & 10.1 & 15.9  &   6644   &  7.0  & 20.6 \\      
    0.99    &   5274    & 5.4 & 12.0 & 4659    & 9.6 &  15.7  &   4416   &  6.2  & 20.0 \\
    \hline
    \end{tabular}
    \caption{Variation of performance of SPIN and OCHMR with respect to confidence score across datasets using Faster-RCNN bounding boxes.}
    \label{tab:supplementary:detector_conf}
\end{table}

  \begin{figure*}
 \captionsetup{font=small}
 \begin{center}

\includegraphics[width=0.75\linewidth, height=0.34\linewidth]{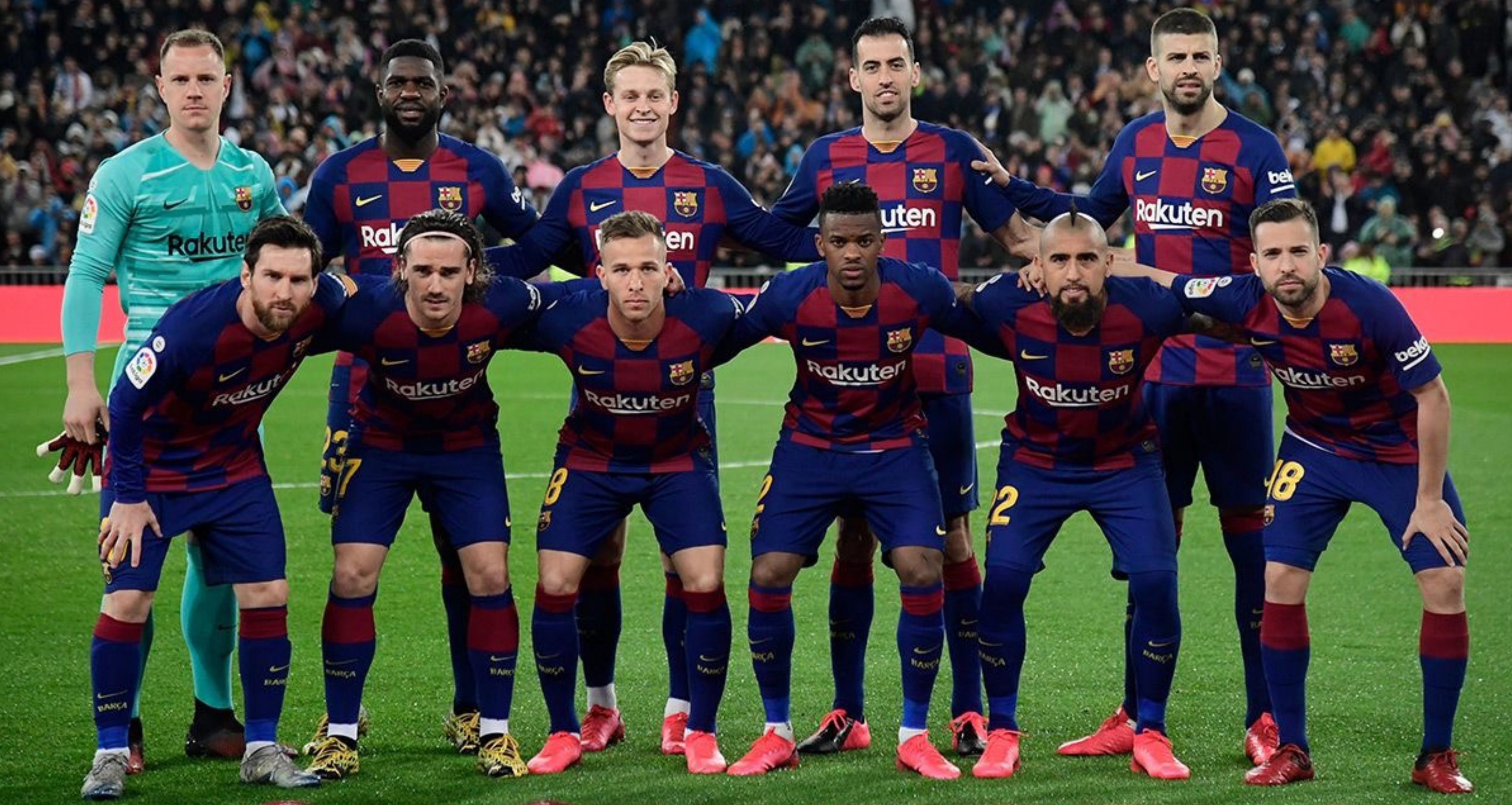}
\includegraphics[width=0.75\linewidth, height=0.34\linewidth]{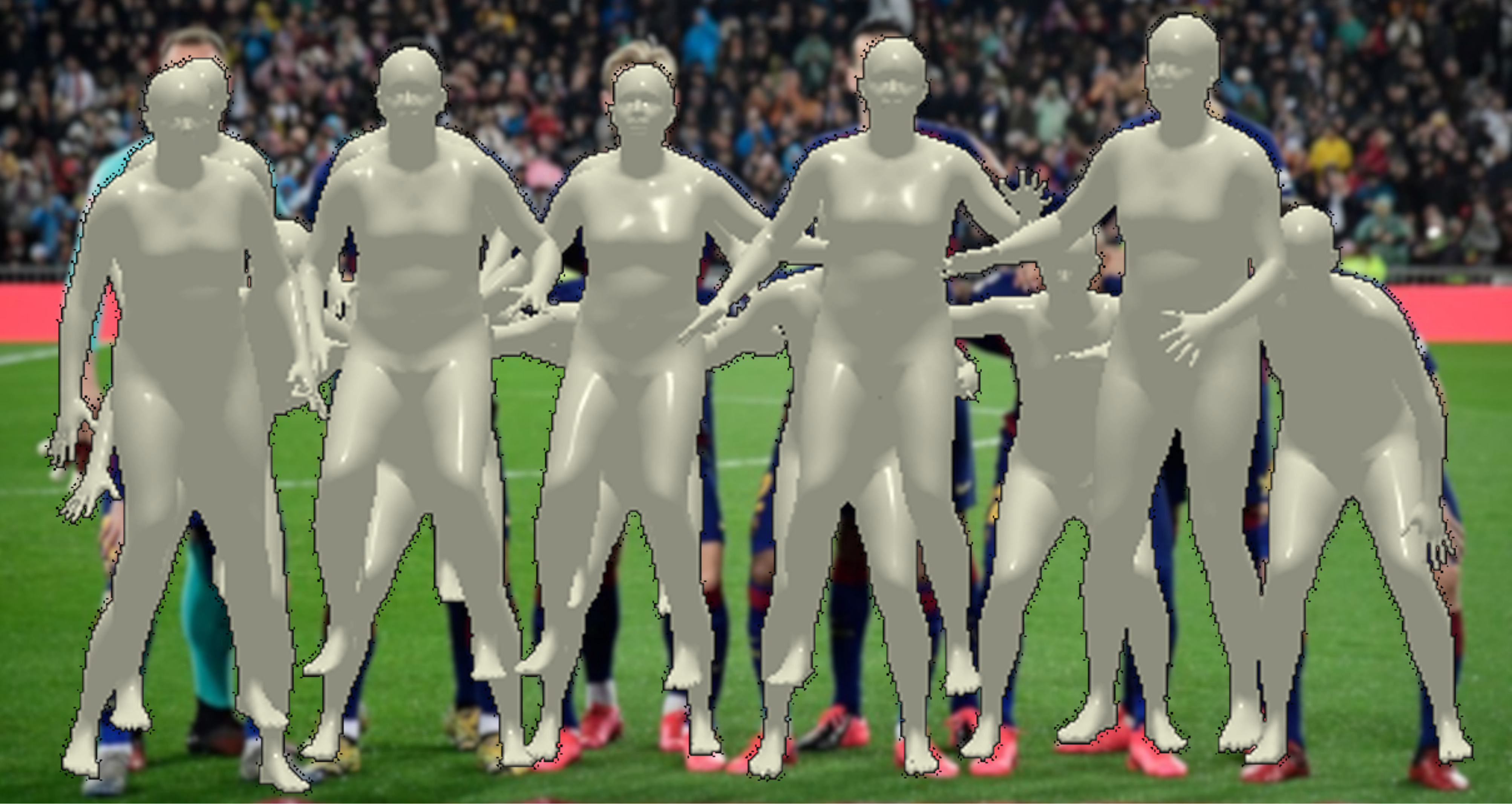}
\includegraphics[width=0.75\linewidth, height=0.34\linewidth]{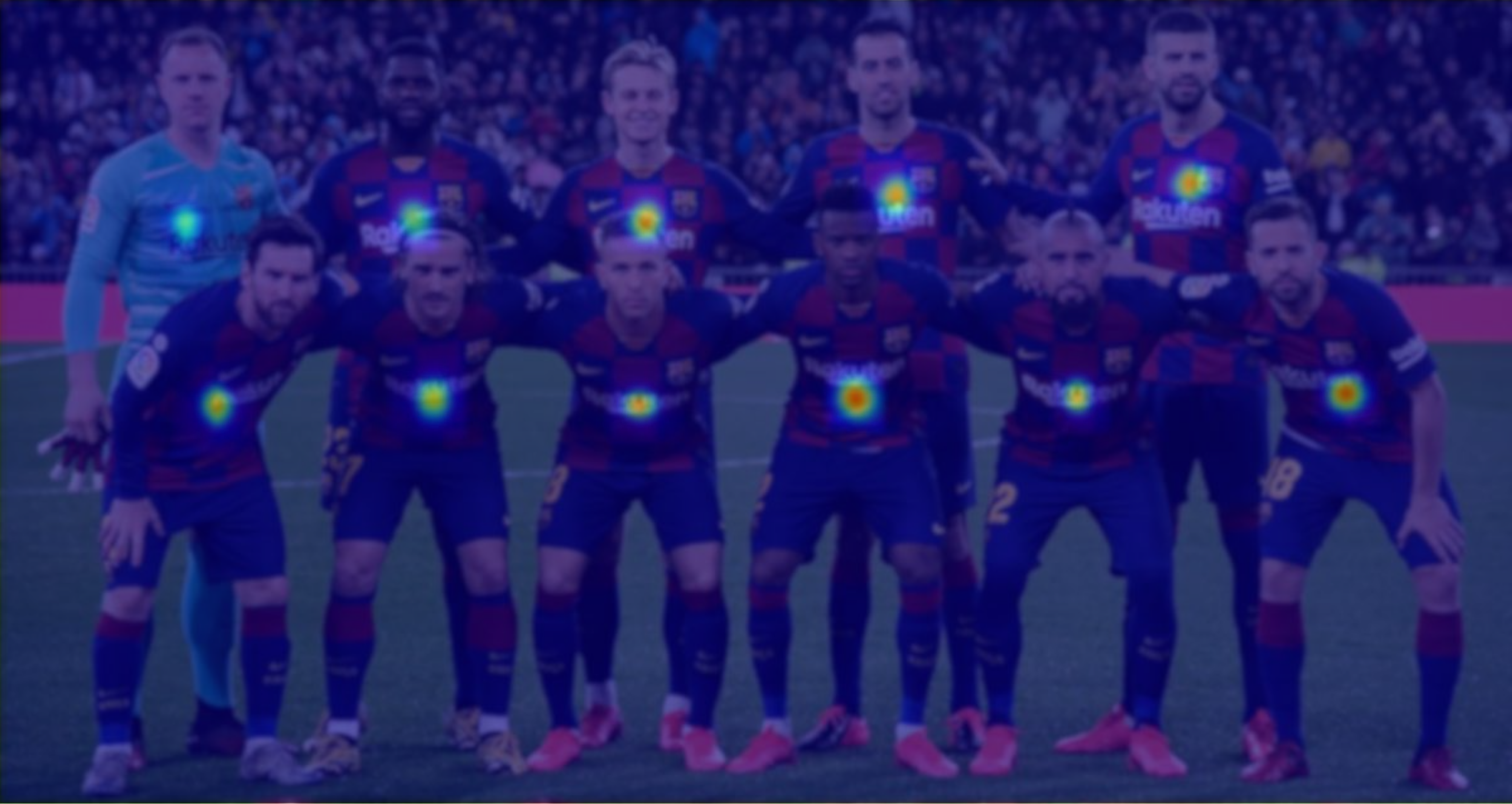}
\includegraphics[width=0.75\linewidth, height=0.34\linewidth]{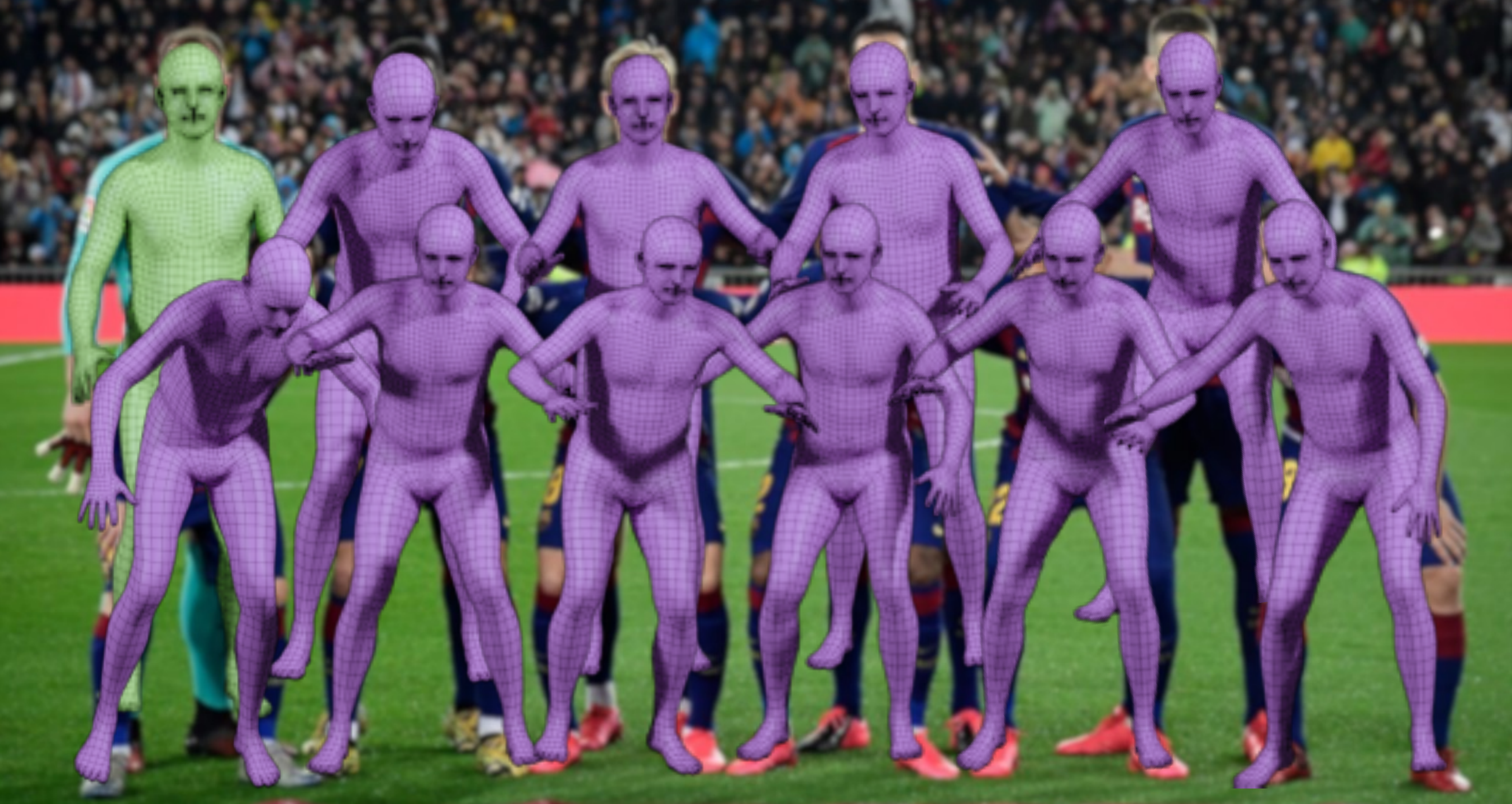}

\end{center}
 \vspace*{-0.2in}
\caption{Qualitative results under challenging depth-ordering scenarios. Each image (top to bottom) shows RGB image, ROMP~\cite{sun2021monocular} predictions, OCHMR global-centermap predictions and OCHMR mesh predictions.}
\label{figure:supplementary:qual_crowd}
 \end{figure*}

 \begin{figure*}
 \captionsetup{font=small}
 \begin{center}

\includegraphics[width=0.95\linewidth]{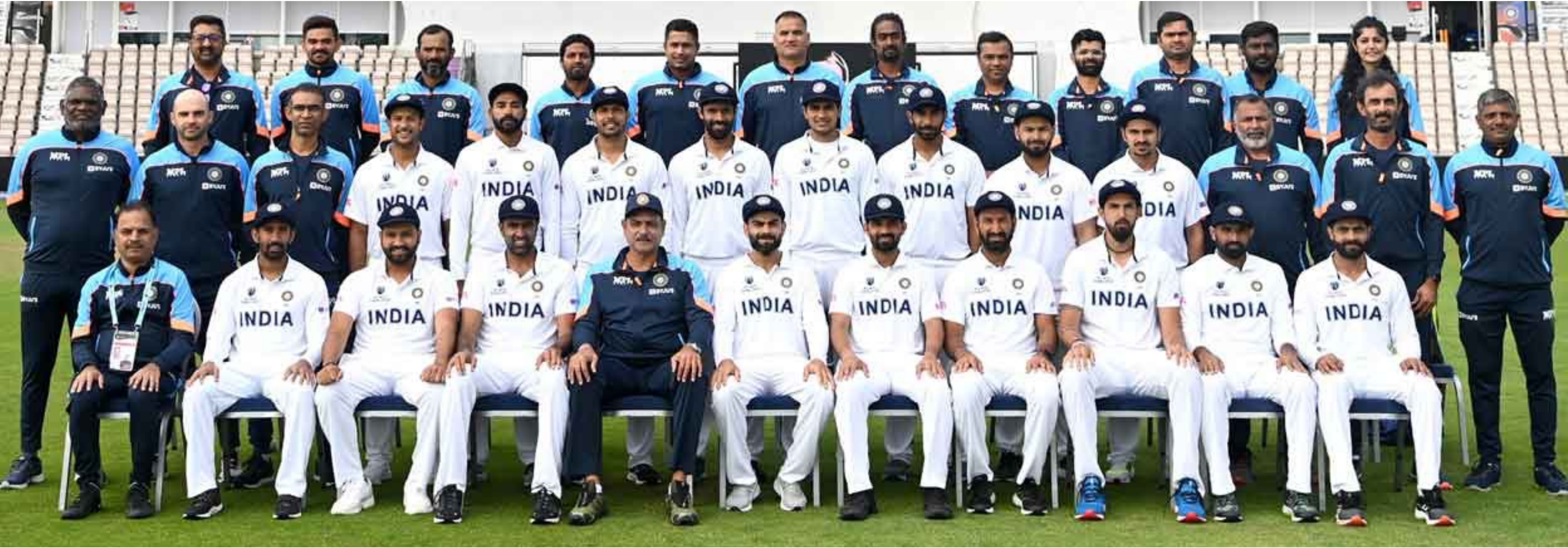}
\includegraphics[width=0.95\linewidth]{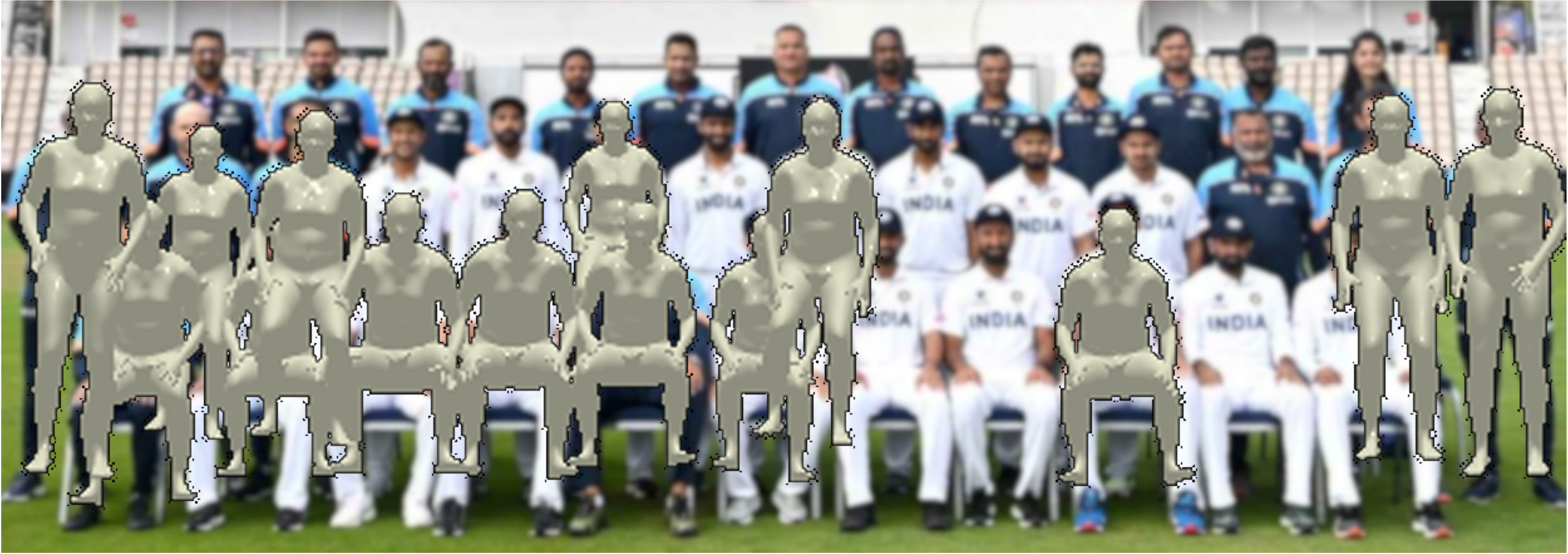}
\includegraphics[width=0.95\linewidth]{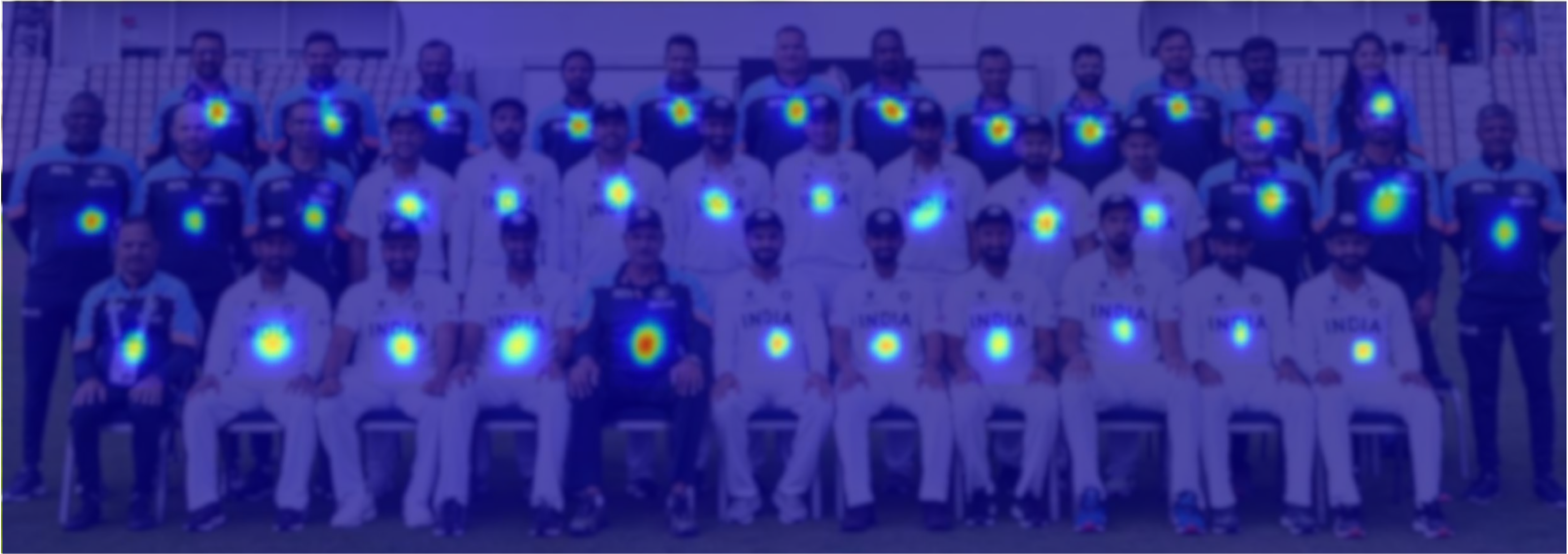}
\includegraphics[width=0.95\linewidth]{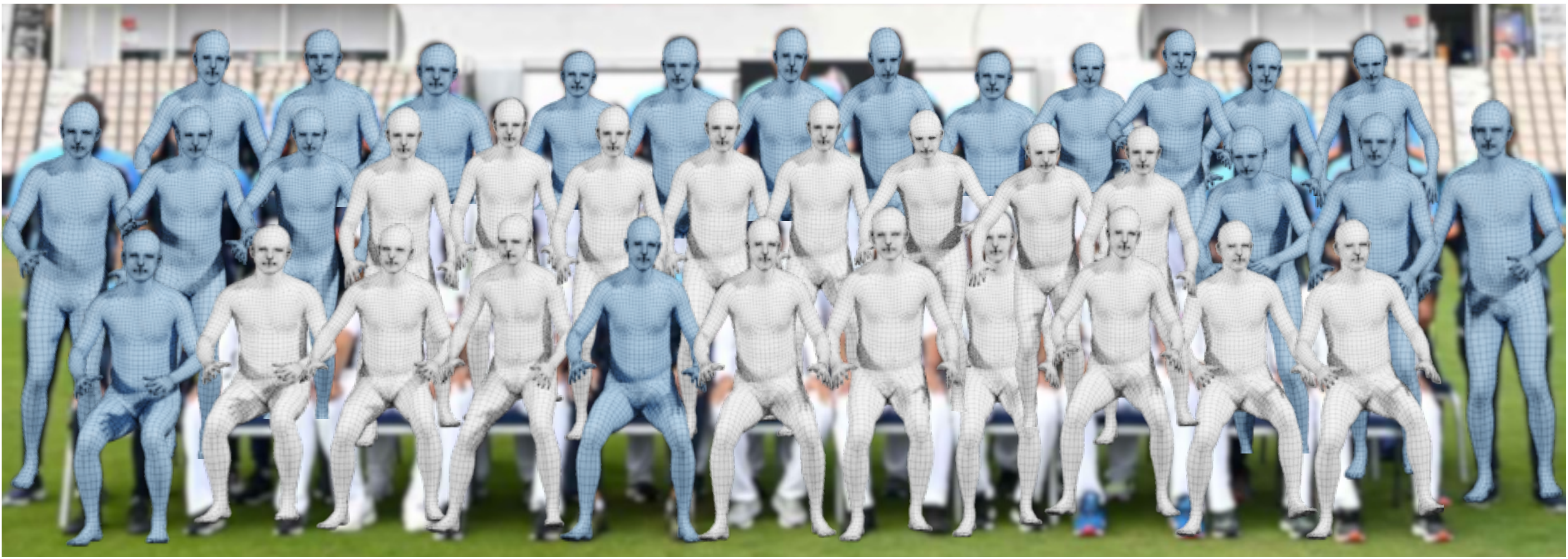}

\end{center}
 \vspace*{-0.2in}
\caption{Qualitative results under severe crowding. Each image (top to bottom) shows RGB image, ROMP~\cite{sun2021monocular} predictions, OCHMR global-centermap predictions and OCHMR mesh predictions. ROMP fails to recover mesh for prominent humans in the image even after setting the confidence threshold to as low as $0.15$. OCHMR recovers accurate meshes for all humans in the image.}
\label{figure:supplementary:qual_crowd}
 \end{figure*}

 \begin{figure*}
 \captionsetup{font=small}
 \begin{center}

\includegraphics[height=0.17\textheight,width=0.48\linewidth]{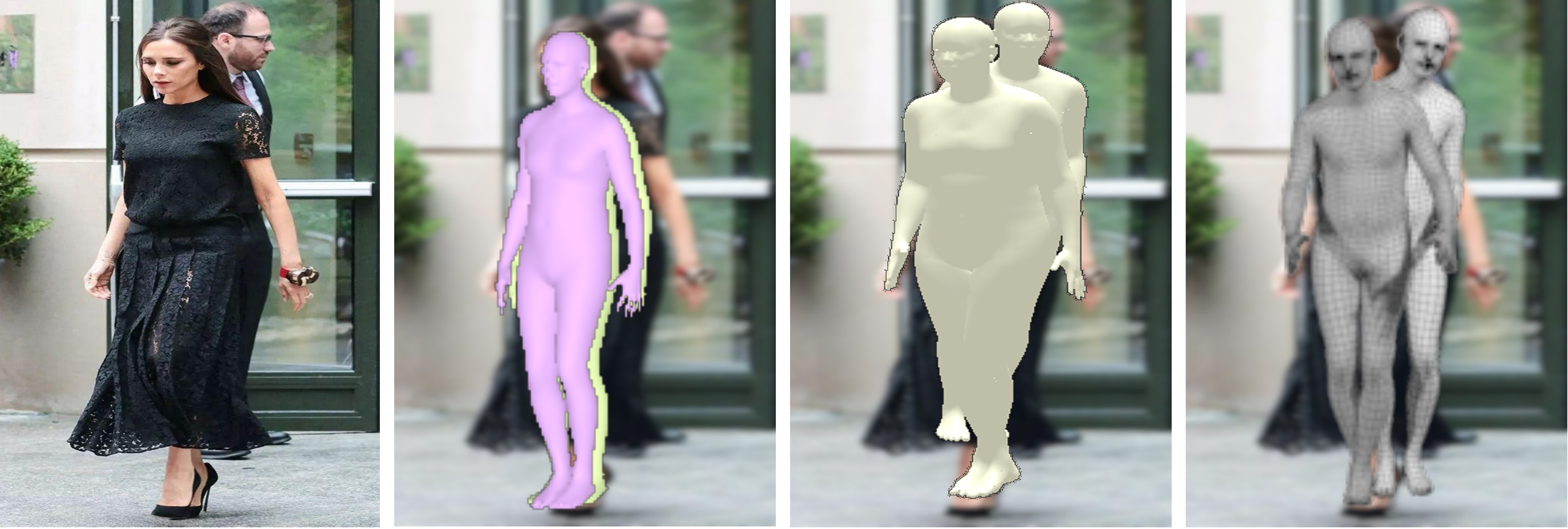}
\includegraphics[height=0.17\textheight,width=0.48\linewidth]{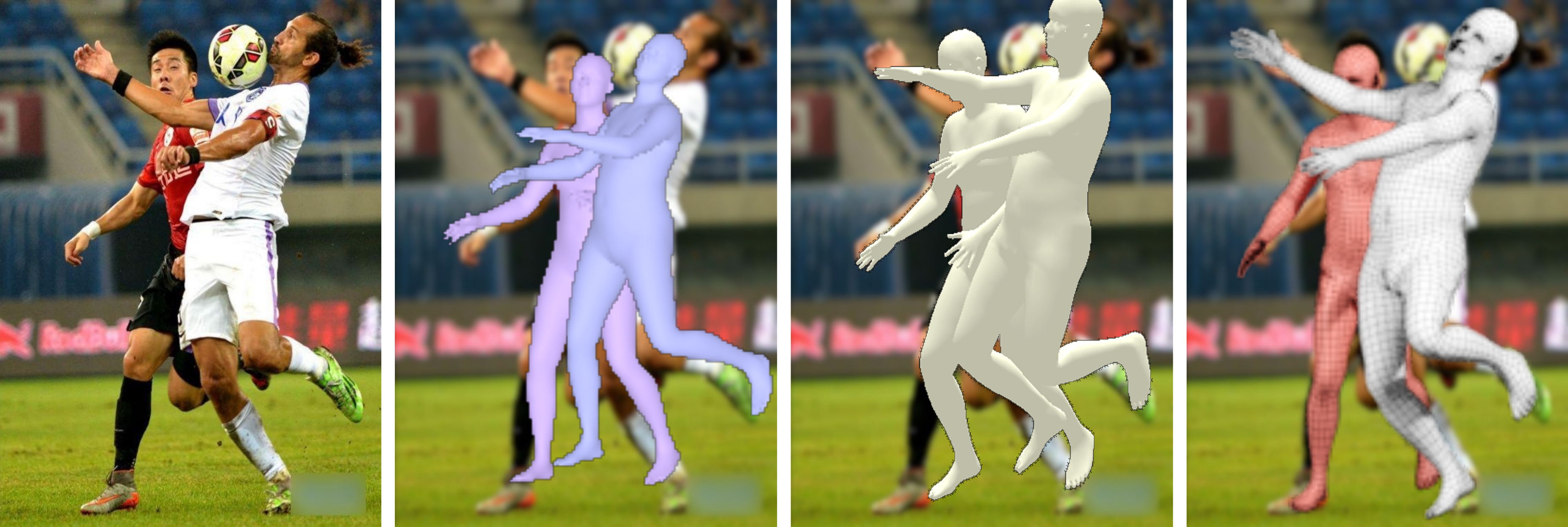}

\includegraphics[height=0.17\textheight,width=0.48\linewidth]{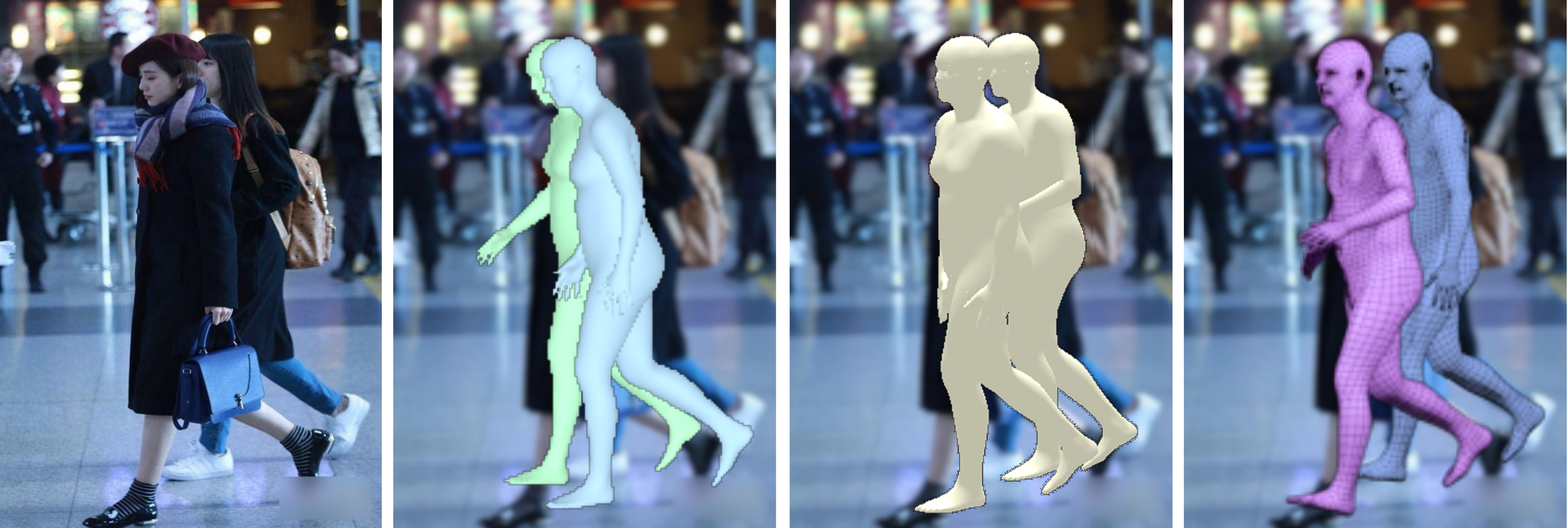}
\includegraphics[height=0.17\textheight,width=0.48\linewidth]{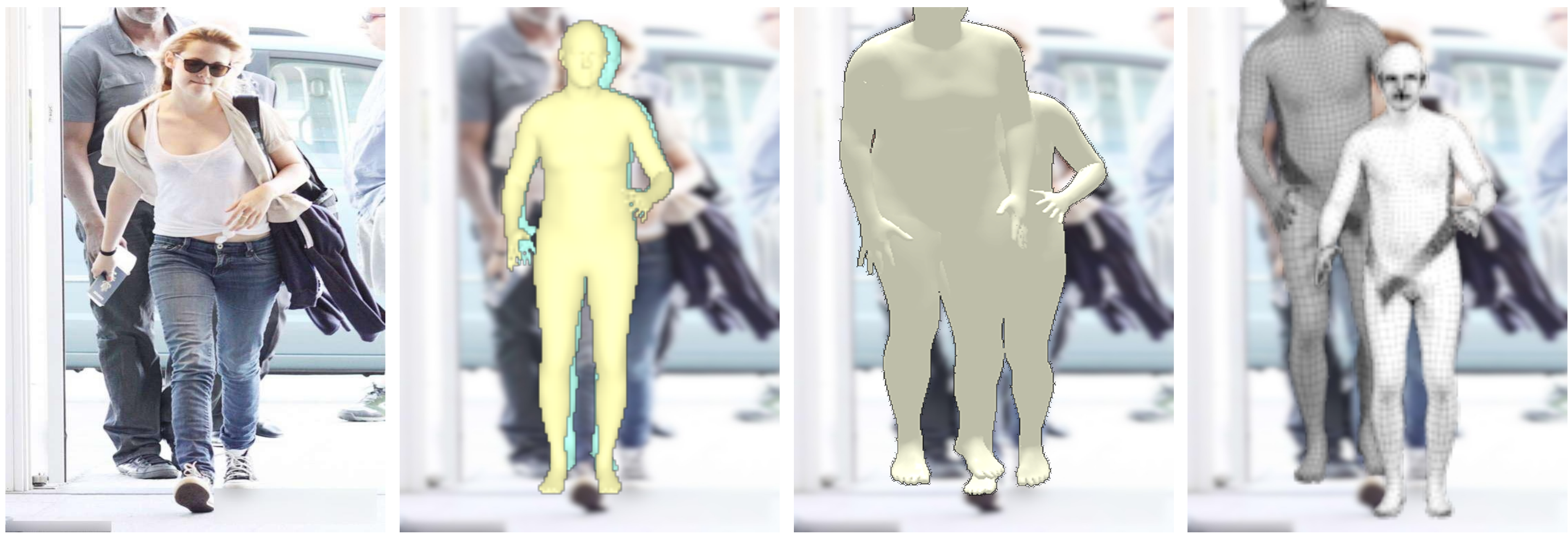}

\includegraphics[height=0.17\textheight,width=0.48\linewidth]{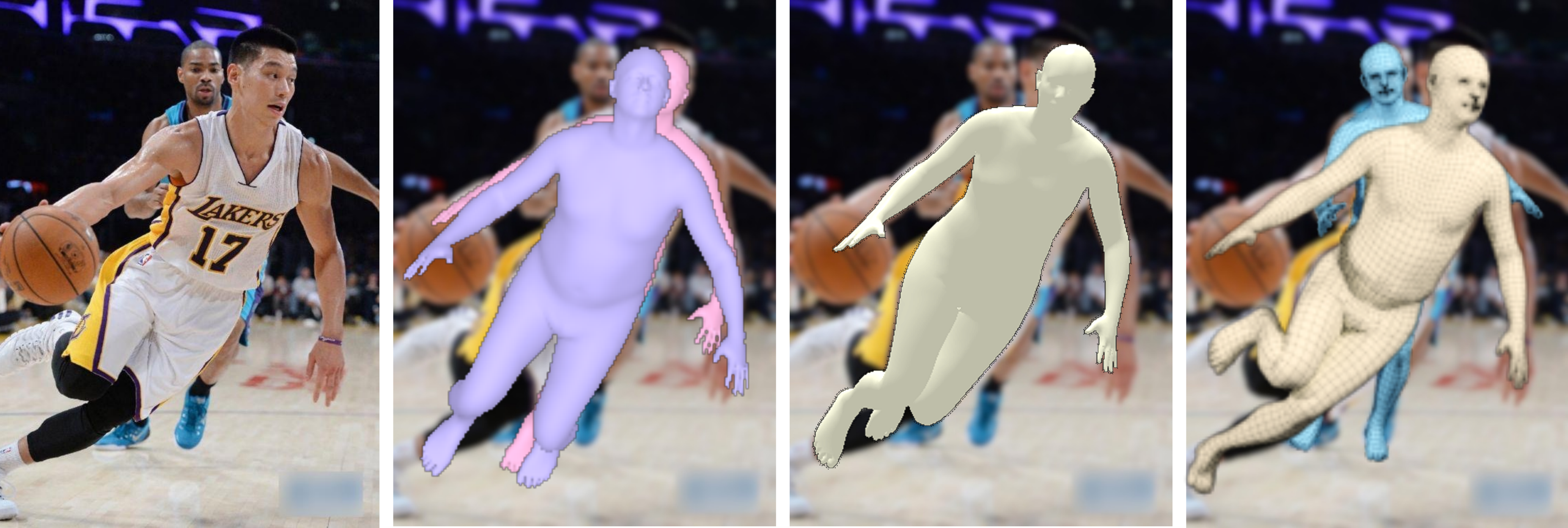}
\includegraphics[height=0.17\textheight,width=0.48\linewidth]{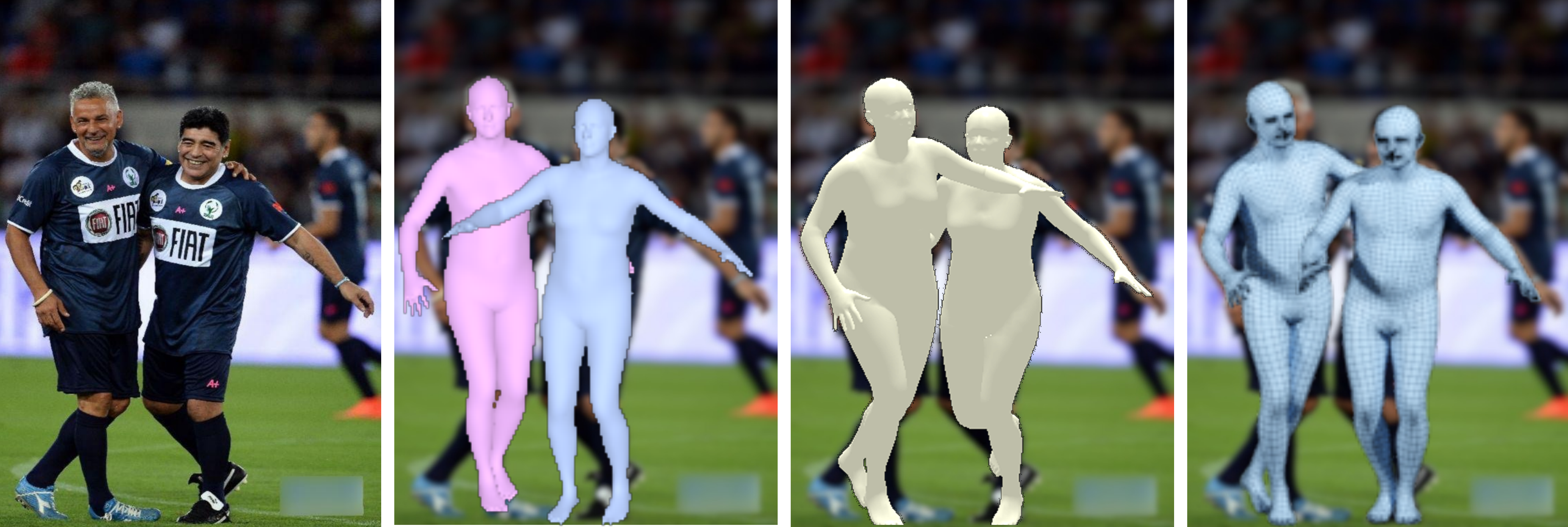}

\includegraphics[height=0.17\textheight,width=0.48\linewidth]{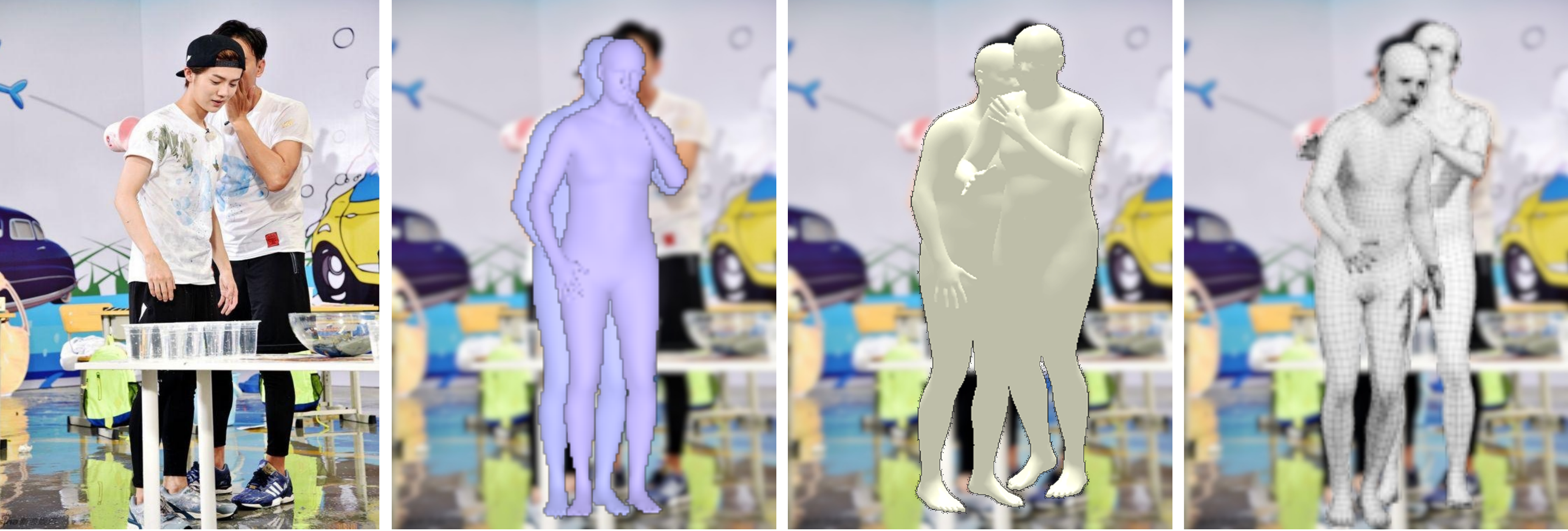}
\includegraphics[height=0.17\textheight,width=0.48\linewidth]{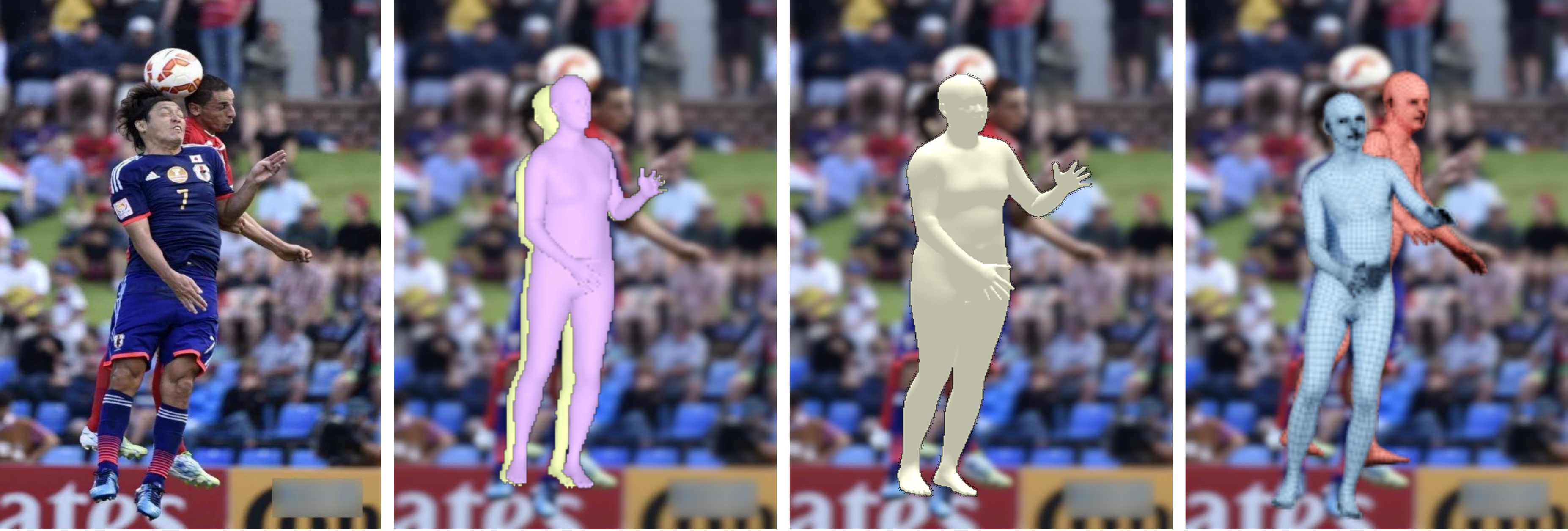}

\includegraphics[height=0.17\textheight,width=0.48\linewidth]{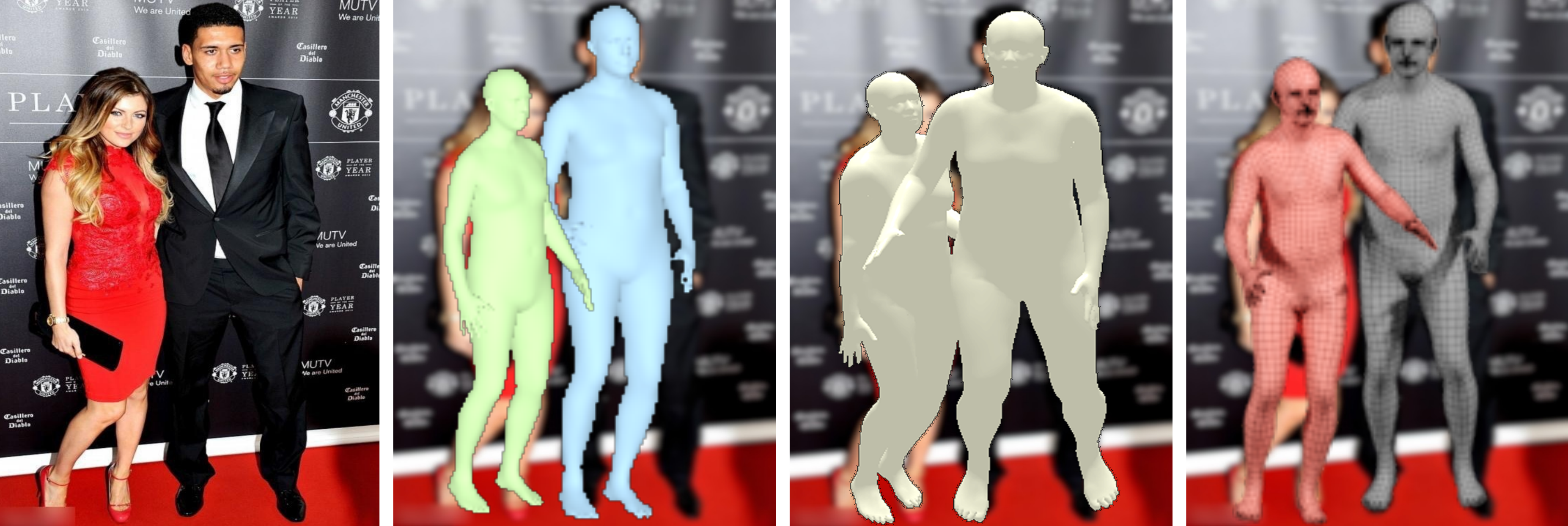}
\includegraphics[height=0.17\textheight,width=0.48\linewidth]{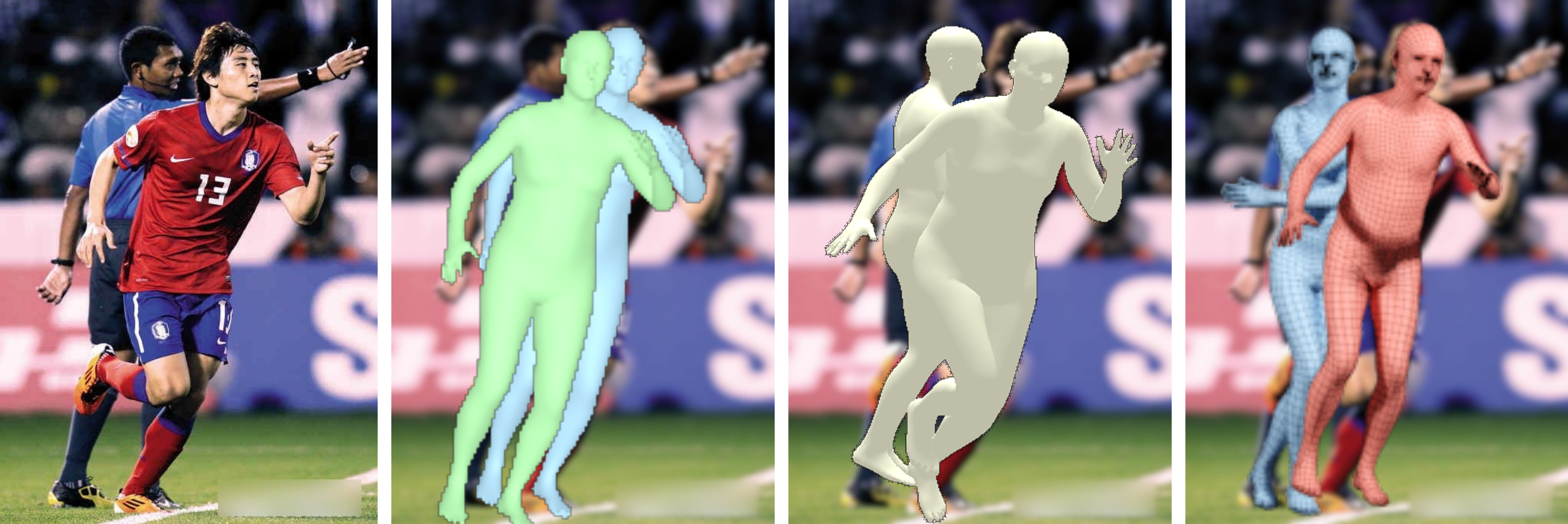}

\includegraphics[height=0.17\textheight,width=0.48\linewidth]{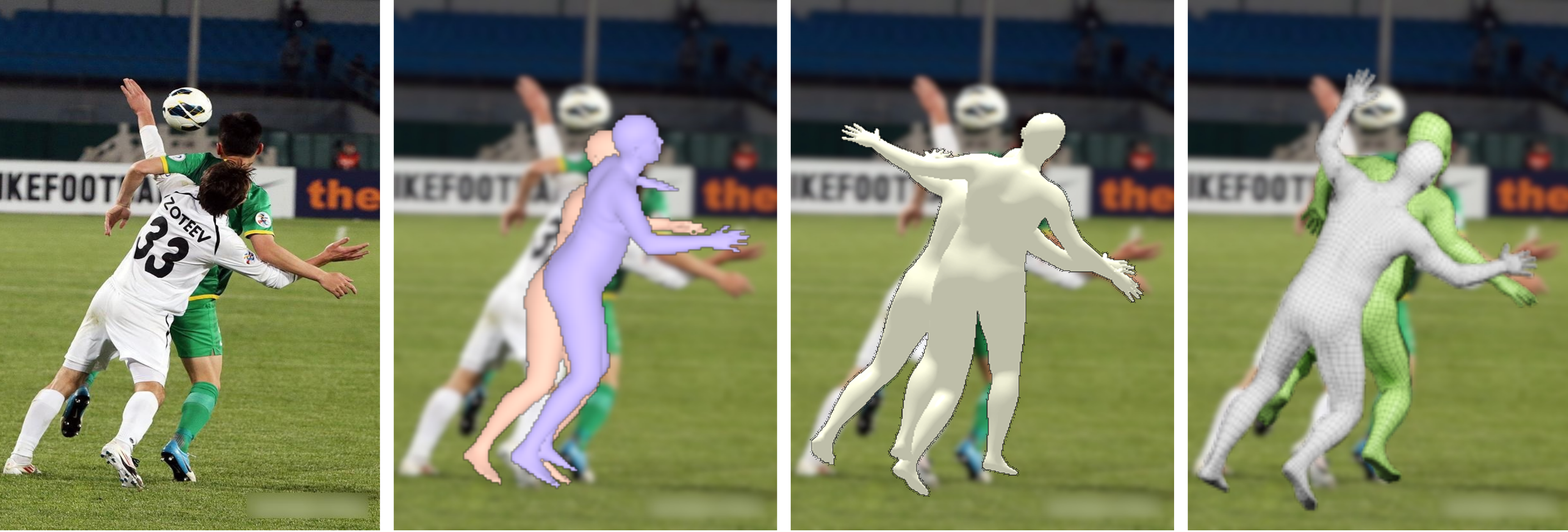}
\includegraphics[height=0.17\textheight,width=0.48\linewidth]{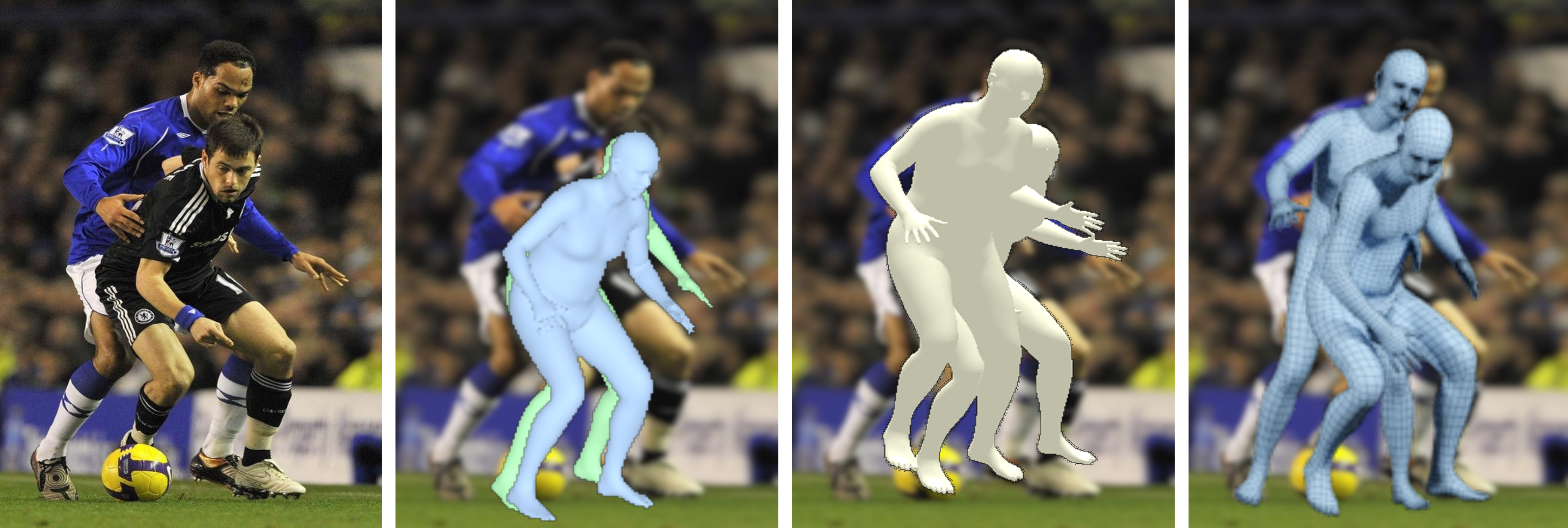}

\end{center}
 \vspace*{-0.2in}
\caption{More qualitative results on the OCHuman dataset. Each image (left to right) shows RGB image, SPIN~\cite{kolotouros2019learning} predictions, ROMP~\cite{sun2021monocular} predictions and OCHMR predictions. OCHMR outputs pose consistent meshes under severe person-person occlusion.}
\label{figure:supplementary:qualitative1}
 \end{figure*}
 
\newpage

\clearpage
\newpage
{\small
\bibliographystyle{ieee_fullname}
\bibliography{references}
}